\documentclass[sigconf]{acmart}
\usepackage{booktabs}
\usepackage{multirow}
\usepackage{colortbl}
\usepackage{xcolor}
\usepackage{graphicx} 
\usepackage{float} 
\definecolor{mycolor}{rgb}{0.74, 0.83, 0.9}
\AtBeginDocument{%
  }

\copyrightyear{2026}
\acmYear{2026}
\setcopyright{cc}
\setcctype{by}
\acmDOI{10.1145/3770855.3817460}

\acmConference[KDD 2026] {Proceedings of the 32nd ACM SIGKDD Conference on Knowledge Discovery and Data Mining V.2}{August 9--13, 2026}{Jeju Island, Republic of Korea.}

\acmBooktitle{Proceedings of the 32nd ACM SIGKDD Conference on Knowledge Discovery and Data Mining V.2 (KDD 2026), August 9--13, 2026, Jeju Island, Republic of Korea}
\acmISBN{979-8-4007-2259-2/2026/08}

\acmSubmissionID{v2dtb031}

\begin{document}

\title{BacktestBench: Benchmarking Large Language Models for Automated Quantitative Strategy Backtesting}


\author{Zhensheng Wang}
\orcid{0009-0003-0395-8572}
\affiliation{%
  \institution{Beijing Normal University}
  \city{Beijing}
  \country{China}
}
\email{jensenwang@mail.bnu.edu.cn}

\author{Wenmian Yang}
\authornote{Corresponding authors.}
\orcid{0000-0001-8493-4449}
\affiliation{%
  \institution{Beijing Normal University}
  \city{Zhuhai}
  \country{China}}
\email{wenmianyang@bnu.edu.cn}

\author{Qingtai Wu}
\orcid{0009-0009-7047-5689}
\affiliation{%
  \institution{Beijing Normal University}
  \city{Zhuhai}
  \country{China}}
\email{qingtaiwu@mail.bnu.edu.cn}

\author{Lequan Ma}
\orcid{0009-0007-8238-0295}
\affiliation{%
 \institution{Beijing Normal University}
 \city{Zhuhai}
  \country{China}}
\email{lequanma@mail.bnu.edu.cn}

\author{Yiquan Zhang}
\orcid{0000-0003-0505-8271}
\affiliation{%
  \institution{Elmleaf Ltd.}
  \city{Shanghai}
  \country{China}}
\email{zhangyq987@hotmail.com}

\author{Weijia Jia}
\orcid{0000-0003-1000-3937}
\authornotemark[1] 
\affiliation{%
  \institution{Beijing Normal University}
  \city{Zhuhai}
  \country{China}}
\email{jiawj@bnu.edu.cn}

\renewcommand{\shortauthors}{Zhensheng Wang, Wenmian Yang, Qingtai Wu, Lequan Ma, Yiquan Zhang, and Weijia Jia}

\begin{abstract}

Quantitative backtesting is essential for evaluating trading strategies but remains hampered by high technical barriers and limited scalability. While Large Language Models (LLMs) offer a transformative path to automate this complex, interdisciplinary workflow through advanced code generation, tool usage, and agentic planning, the practical realization is significantly challenged by the current lack of a large-scale benchmark dedicated to automated quantitative backtesting, which hinders progress in this field. To bridge this critical gap, we introduce BacktestBench, the first large-scale benchmark for automated quantitative backtesting. Built from over 6 million real market records, it comprises 18,246 meticulously annotated question-answering pairs across four task categories: metrics calculation, ticker selection, strategy selection, and parameter confirmation. We also propose AutoBacktest, a robust multi-agent baseline that translates natural language strategies into reproducible backtests by coordinating a Summarizer for semantic factor extraction, a Retriever for validated SQL generation, and a Coder for Python backtesting implementation. Our evaluation on 23 mainstream LLMs, complemented by targeted ablations, identifies key factors that influence end-to-end performance and highlights the importance of grounded verification and standardized indicator representations. The dataset and code will be publicly released at \url{https://github.com/jensenw1/BacktestBench}.

\end{abstract}

\ccsdesc[500]{Computing methodologies~Language resources}
\ccsdesc[500]{Information systems~Question answering}
\ccsdesc[500]{Computing methodologies~Multi-agent systems}

\keywords{Large Language Models, Question Answering, Quantitative Investing, Automated Backtesting, Benchmark, Multi-agent Framework}


\maketitle

\section{Introduction}

\begin{figure}[t]
 \centering
 \includegraphics[width=\columnwidth]{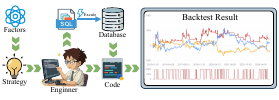}
 \Description{}
 \centering
 \caption{Real-world workflow of a quantitative backtesting engineer.}
 \label{fig:Real-world_workflow}
\end{figure}

\noindent\emph{“History does not repeat itself, but it often rhymes.”} \\
\hspace*{\fill} --- Mark Twain


In quantitative investing, this aphorism captures the fundamental role of backtesting: it involves replaying trading logic on extensive historical data to rigorously assess strategy robustness and its potential adaptability to future market regimes~\cite{pardo2011evaluation}. By reporting quantitative metrics such as the sharpe ratio~\cite{sharpe1998sharpe} and maximum drawdown~\cite{magdon2004maximum}, backtesting provides an objective evaluation framework, significantly reducing reliance on subjective judgment and enhancing comparability across diverse investment strategies~\cite{de2018advances}. However, the traditional backtesting workflow, as illustrated in Figure~\ref{fig:Real-world_workflow}, is inherently complex and interdisciplinary, requiring specialized expertise. Strategy engineers must precisely construct factor combinations, issue accurate historical data queries for specific tickers and time windows, and implement error-free code to execute intricate strategy logic. This process is constrained by high technical barriers, demanding meticulous data retrieval, strict adherence to market timing, and precise trading rule implementation.


The recent advent of Large Language Models (LLMs) and their advanced capabilities in code generation, logical reasoning~\cite{DBLP:journals/corr/abs-2107-03374,DBLP:journals/corr/abs-2308-12950,DBLP:conf/iclr/LuoX0SGHT0LJ24}, and, more importantly, agentic planning~\cite{DBLP:conf/iclr/YaoZYDSN023}, tool invocation~\cite{toolformer}, and iterative refinement~\cite{Reflexion} in interactive environments, presents a transformative opportunity for automating quantitative backtesting. Faced with an exploding search space of factor combinations and rapidly shifting market demands, manual strategy implementation and verification processes can no longer meet the efficiency and scalability requirements of modern quantitative research. This context prompts us to investigate whether LLM-based agents can reliably automate the entire backtesting pipeline, encompassing data retrieval, strategy translation, and execution. Such an endeavor also serves as a practical touchstone for evaluating the capability of Artificial General Intelligence (AGI)~\cite{DBLP:books/ox/90/Turing90} on complex vertical-domain problems.

While the prospect of natural language-driven end-to-end automated backtesting is highly promising, its practical realization presents significant challenges. It demands that models simultaneously possess precise semantic understanding of financial strategies, robust structured data retrieval capabilities, and highly accurate code generation skills~\cite{DBLP:conf/nips/XieHZLPLH23}. Critically, the academic community currently lacks high-quality evaluation benchmarks specifically designed for such complex financial reasoning tasks. Existing code generation or Text-to-SQL datasets~\cite{DBLP:conf/nips/LiHQYLLWQGHZ0LC23,DBLP:conf/emnlp/YuZYYWLMLYRZR18,DBLP:conf/iclr/LeiCYCSSSGHYZX025}, while valuable, do not adequately capture the unique temporal logic and multi-step reasoning requirements inherent in quantitative backtesting.

To bridge this critical gap, we introduce BacktestBench, a novel large-scale benchmark dataset tailored for the field of automated backtesting. Constructed from over 6 million real historical market records from China's three major exchanges, BacktestBench comprises 18,246 high-quality question-answering (QA) pairs. Each sample is meticulously annotated, covering the entire information chain from a natural language strategy description to core factor extraction, SQL data query generation, Python backtesting code implementation, and the final objective answer. Furthermore, to comprehensively evaluate the holistic capabilities of models, this benchmark systematically designs four distinct task categories: metrics calculation, ticker selection, strategy selection, and parameter confirmation. These categories fully encompass real-world backtesting scenarios such as strategy comparison, optimization, and asset allocation, thereby providing a rigorous yardstick for measuring the intelligence level of LLMs in this vertical domain.

To systematically address BacktestBench's challenges and effectively evaluate LLMs' reasoning and execution capabilities in this domain, this study proposes AutoBacktest, a multi-agent collaborative system serving as a robust baseline. It transforms unstructured natural language strategy descriptions into reproducible backtest results, mimicking human quantitative researchers' workflows by decoupling complex decision chains into specialized sub-tasks. Specifically, AutoBacktest coordinates three functionally specialized agents: (1) the Summarizer, responsible for semantic-level extraction of financial indicators (including both trading factors and key performance indicators); (2) the Retriever, which handles data-level precise querying and quality verification; and (3) the Coder, focusing on logic-level code implementation and backtest execution. This layered, modular architecture not only significantly reduces the inherent difficulty of end-to-end generation but also enhances the robustness of the entire backtesting pipeline through self-verification mechanisms at each stage, thereby establishing a standardized reference baseline for future research.

The main contributions of this paper are as follows:

\begin{itemize}
    \item We establish BacktestBench, a large-scale benchmark for automated quantitative backtesting. Built on over 6 million real market records, it covers metrics calculation, ticker selection, strategy selection, and parameter confirmation.

    \item We propose AutoBacktest, a robust multi-agent baseline system that translates natural language strategies into executable backtests via semantic factor extraction, SQL generation, and Python code execution.

    \item We evaluate 23 mainstream LLMs on BacktestBench and perform ablation studies to identify critical factors influencing end-to-end backtesting performance.
\end{itemize}

\section{Dataset Construction}

\subsection{Data Source}


To construct a backtesting dataset tailored to real-world scenarios, we perform rigorous cleaning and integration of raw stock data provided by Elmleaf Information Technology. First, we collect daily trading data for all stocks listed on the Shanghai, Shenzhen, and Beijing Stock Exchanges from January 2, 2020, to September 30, 2025. Post-cleaning, the dataset comprises 6,549,254 records covering 5,401 listed companies over a duration of 1,395 trading days. More detailed schema descriptions and additional statistics are deferred to Appendix~\ref{appendix:Database}.


\subsection{Core Concepts}

In Quantitative Investing, we distinguish four closely related concepts: factor, signal, strategy, and KPI.

A \emph{factor} is a numerical variable computed from raw market observations (e.g., open, high, low, close, volume) via a deterministic statistical or technical formula, and it quantifies a specific market pattern or state~\cite{DBLP:conf/kdd/YuX0PHTH23}.
For example, the 5-day moving average (MA5) maps a price series to a scalar time series that represents the average price over the past five trading days.

A \emph{signal} is a Boolean decision derived by applying a constraint or threshold rule to one or more factors at time $t$, indicating whether a trading condition is satisfied.
For instance, a buy signal can be defined as $\texttt{Open}_t > \texttt{MA5}_{t-1}$, where the factor value uses only information available before time $t$ to avoid look-ahead bias.
Similarly, a sell signal can be defined by an opposite inequality or a different constraint on another factor.

A \emph{strategy} specifies how to combine buy and sell signals into executable trading actions under a backtesting protocol, including the entry and exit logic, position state transitions, and required constraints (e.g., long-only, alternating buy and sell).
For example, a simple long-only strategy enters when the buy signal fires, exits when the sell signal fires, and holds the position otherwise.

A \emph{KPI} (key performance indicator) is a quantitative metric used to evaluate factors, signals, and strategies under a backtesting protocol, covering predictability, risk, and implementability.

In summary, factors provide the numeric basis, signals convert factors into actionable conditions, strategies compose buy and sell signals into a complete and reproducible trading policy for backtesting, and KPIs provide the evaluation criteria that connect these concepts to measurable and comparable outcomes.

\subsection{Factor and KPI Selection}

\subsubsection{Factor Selection}

Facing a massive number of candidate factors, this study adheres to two core screening principles:
\begin{itemize}
    \item \textbf{Data Availability:} The existing database schema must fully support the foundational fields required for factor calculation.
    \item \textbf{Computational Determinism:} Factors must possess strict mathematical formulations to ensure the uniqueness and reproducibility of calculation results. Based on this principle, we exclude factors involving randomness or subjective judgment, such as news sentiment scores generated by LLMs based on non-deterministic seeds, or soft metrics relying on manual annotation.
\end{itemize}

Based on these criteria, we select 43 factors spanning Risk, Quality, Momentum and Technical categories, including 4 sell-only, 2 buy-only and 37 dual-use factors. A detailed list of all factors is provided in Appendix \ref{appendix:Factors}.

\subsubsection{KPI Selection}
Guided by domain experts, we select seven KPIs as the primary evaluation criteria: Return Ratio, Maximum Drawdown, Volatility, Annual Sharpe Ratio, Win Rate, Profit Loss Ratio, and Calmar Ratio \cite{eling2007does}. The Calmar Ratio is defined as the ratio of annualized return to maximum drawdown and serves as a downside-risk-sensitive indicator.
Detailed definitions of all seven KPIs are provided in Appendix Table~\ref{tab:all_KPI_used}.

\subsection{Function Implementation}

\subsubsection{Implementation of Atomic Strategy Functions from Factors.}
To transform 43 theoretical factors into executable engineering objects, we encapsulate them into 80 atomic strategy functions using Pandas and NumPy. Each function couples factor computation with explicit signal-triggering logic, with 41 dedicated to sell-side and 39 to buy-side operations. These collectively form our signal pool.




Each atomic strategy function takes the target ticker's historical data and specific hyperparameters (e.g., lookback window, decision threshold) as inputs. These functions encapsulate both factor computation and signal evaluation, returning a list of trading dates where the strategy is triggered. The \textit{threshold} parameter is specifically used to define the boundary conditions for the logic, such as determining whether a value deviation is significant enough to generate a signal. To guarantee realistic backtesting, we strictly avoid look-ahead bias by ensuring that the decision to trade at day \( T \) relies exclusively on statistics computed from information available up to \( T-1 \).
An illustrative example is provided in the appendix \ref{appendix:Factors}.

\subsubsection{Implementation of KPI Computation Functions}

Because several financial KPIs are sensitive to implementation choices and LLMs can hallucinate when reasoning about these computations, we adopt a standardized backtesting protocol with deterministic trading rules and unambiguous metric definitions to ensure unique and reproducible results.

The protocol fixes the execution microstructure to a long-only setup with strictly alternating buy and sell operations, disallows intraday round trips (T+0), executes buys at the opening price and sells at the closing price, and forcibly liquidates any remaining position at the end of the backtest window.

We simulate a fixed-capital portfolio without leverage, enforce round-lot trading (at least 100 shares), and set transaction costs and taxes to zero in the baseline experiments.
We evaluate performance under a daily mark-to-market accounting convention with annualization based on 252 trading days per year and a constant daily risk-free rate of 0.0001 for the Annual Sharpe Ratio, and we assume the input time series is complete and strictly ordered by trading date.
We provide the full specification of the execution mechanism, position sizing constraints, return computation, and data integrity assumptions in Appendix~\ref{appendix:BacktestingProtocol}.

The backtesting logic described above is implemented in Python and encapsulated into a dedicated metrics calculation function. This function takes as input a \texttt{DataFrame} containing market data together with the series of buy and sell signals generated by the signal pool, and returns the seven KPI values defined in the evaluation framework.

To ensure both correctness and robustness of the implementation, this study applies a double-blind verification procedure. Two quantitative engineering experts independently implement the same metric computation function strictly following the shared backtesting protocol, and then conduct cross-validation on ten randomly selected market data samples. Only when the outputs of the two implementations match exactly across all test cases is the implementation accepted as correct; the preferred version is then chosen as the final backtesting engine after discussion. This procedure effectively mitigates potential logical errors at the code level and enhances the reliability of the experimental evaluation toolkit.

\subsection{Strategy Code Construction}



To construct a diverse corpus for quantitative investing, we design an automated framework that dynamically synthesizes executable backtesting code. 
The framework operates by sampling and combining atomic strategy functions. 
Combinatorial analysis demonstrates that selecting up to four atomic functions without replacement yields a vast search space comprising 92,170 buy strategies and 112,791 sell strategies. 
Leveraging Python reflection, the system instantiates these atomic operators and maps them into four canonical decision tasks: metrics calculation, ticker selection, parameter confirmation, and strategy selection. 
Detailed construction procedures and generation statistics are provided in Appendix~\ref{appendix:DataGenDetails}.

\paragraph{Metrics Calculation.}
This task family evaluates the model's ability to compute quantitative performance indicators for a single strategy under a fixed market environment.
Each instance presents a natural language description of the strategy and a specified KPI (e.g., Sharpe Ratio or Maximum Drawdown), and the model is required to output the corresponding numerical value.

\paragraph{Ticker Selection.}
Ticker selection tasks emulate choosing the best-performing asset from a candidate universe under a shared strategy specification.
Given a fixed strategy and multiple ticker histories, the model must identify the asset that optimizes the target KPI. 

\paragraph{Parameter Confirmation.}
Parameter confirmation tasks focus on selecting hyperparameters within an otherwise fixed strategy architecture.
The model is asked to determine which candidate parameter value (such as a threshold or lookback window) yields the best KPI when applied to the same underlying asset and backtest window.

\paragraph{Strategy Selection.}
Strategy\_selection tasks sit at the top of the decision hierarchy and require choosing among multiple competing strategy logics under identical market conditions.
Given several candidate strategies sharing the same underlying, horizon, initial capital, and evaluation metric, the model must decide which logic achieves the highest performance.

\paragraph{Remarks.}
Across all four task families, each instance is paired with executable Python code, a standardized backtesting environment, and ground-truth labels, enabling end-to-end evaluation from natural language to quantitative outcomes.
Further implementation details, including data sampling rules, code synthesis pipelines, and filtering criteria, are provided in Appendix~\ref{appendix:DataGenDetails}.

\begin{figure}[t]
 \centering
\includegraphics[width=0.95\columnwidth]{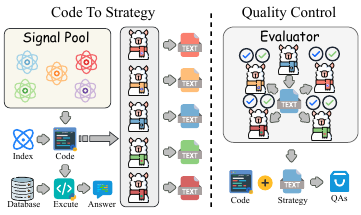}
\Description{}
 \centering
 \caption{Pipeline of natural language strategy generation and evaluation.}
 \label{fig:Code2Strategy}
\end{figure}

\subsection{Strategy Description Generation}

After constructing the corpus of executable strategy code, we adopt a code-to-text reverse-engineering paradigm to obtain human-readable strategy descriptions aligned with actual backtesting behavior.
To ensure both semantic accuracy and financial soundness, we employ a multi-model generation and evaluation pipeline with strict acceptance rules.

\subsubsection{Multi-Model Parallel Generation}

We select five state-of-the-art open-source LLMs as generation agents: Kimi-K2-Thinking-BF16 \cite{DBLP:journals/corr/abs-2507-20534}, MiniMax-M2.1-BF16 \cite{DBLP:journals/corr/abs-2506-13585}, GLM-4.7 \cite{DBLP:journals/corr/abs-2508-06471}, GPT-OSS-120B-BF16 \cite{DBLP:journals/corr/abs-2508-10925}, and Qwen3-235B-A22B-Thinking-2507 \cite{DBLP:journals/corr/abs-2505-09388}.
These models are deployed on-premise using 72 NVIDIA A800-SXM4-80GB GPUs.
For each strategy code sample, each of the five models independently performs reverse parsing and rewriting to produce diverse natural-language descriptions in an instructional style tailored to backtesting scenarios.

Crucially, the input to these models extends beyond raw function bodies.
As illustrated in our codebase, each atomic strategy operator and KPI function is annotated with comprehensive docstrings that explicitly define the \textit{Strategy Logic}, \textit{Mathematical Formula}, and \textit{Input/Output Specifications}.
This rich semantic context provides ground-truth guidance, significantly reducing ambiguity and minimizing the risk of model hallucination during the generation process.
As illustrated on the left side of Figure~\ref{fig:Code2Strategy}, this procedure follows a code-to-strategy pipeline that transforms executable backtesting programs into aligned natural language strategies.
Prompt design details, including how we enforce temporal consistency and separate system-level defaults from strategy-specific logic, are provided in Appendix~\ref{appendix:NL_details}.

\subsubsection{Automated Filtering}
To ensure high corpus quality, we implement a rigorous automated filtering mechanism based on peer review.
For a natural language strategy description generated by a specific model (e.g., model A), the other four models (e.g., models B, C, D, and E) serve as independent auditors.
Each auditor evaluates the code--text pair along two dimensions: \textit{code fidelity} and \textit{strategy validity}.
Code fidelity ensures the text faithfully reconstructs the program logic, while strategy validity verifies that the trading rules are financially logical and practically executable.
We enforce a strict acceptance protocol: a sample is retained if and only if it receives unanimous consensus from all four peer evaluators across both dimensions (totaling eight positive votes).
If multiple descriptions for a single strategy code pass this check, one is randomly selected to prevent duplication.

\subsection{Dataset Statistics and Human Validation}
Applying the rigorous filtering pipeline to the initial pool of 33,003 strategy programs yields a finalized corpus of 18,246 high-quality code--text pairs. 
These instances are stratified by task type into training (10,215), validation (4,195), and testing (3,836) sets. 
Detailed distributions regarding task families and comparative examples of accepted versus rejected descriptions are provided in Appendix~\ref{appendix:NL_details}.

To statistically quantify the quality of the generated corpus, we conduct a human audit on a random sample of 200 instances (representing approximately 5\% of the test set). 
Five volunteers assess the \textit{completeness} of the strategy descriptions, reporting a pass rate of 97.5\% (195/200). 
Simultaneously, six experts in Python implementation verify \textit{code correctness}, confirming a 98.5\% (197/200) consistency between the code and the trading rules.

Furthermore, to benchmark the \textit{realism} of the synthetic data against human standards, we manually construct a separate set of 70 strategies across seven KPI categories. 
This expert-crafted subset is developed through a forward generation process that encompasses natural language description, SQL data retrieval, and backtesting implementation, which is subsequently subjected to rigorous expert cross-validation.

\begin{figure}[t]
 \centering
\includegraphics[width=0.95\columnwidth]{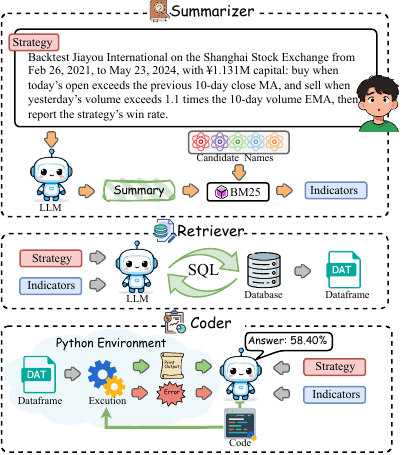}
\Description{}
 \centering
 \caption{Overall framework of the AutoBacktest.}
 \label{fig:Workflow}
\end{figure}

\section{AutoBacktest}

To address the challenges of automated backtesting posed by this dataset, we design a multi-agent framework that mimics the workflow of a quantitative researcher. Three specialized agents cooperate in a pipeline: the Summarizer parses natural language strategies into structured indicator representations, the Retriever generates and validates executable SQL to fetch historical market data, and the Coder produces and runs backtesting code to obtain final numerical results. A high-level description is given below, while interaction details, prompts, and implementation code are deferred to Appendix~\ref{appendix:AgentFlowAndCode}.

\subsection{Summarizer}
The Summarizer first employs an LLM, guided by predefined prompts, to extract keywords related to factors and KPIs from the user's natural language strategy. These summarized keywords then query a BM25 retrieval mechanism, which matches them against a comprehensive library to identify precise standard indicator names. The final output is a structured list of these verified standard indicator names, stored in the shared intermediate state. The concrete prompt design, JSON schema, and reference implementation are provided in Appendix~\ref{appendix:AgentFlowAndCode}.


\subsection{Retriever}
The Retriever builds on the Summarizer's output to construct the data layer for backtesting. 
It first maps the identified indicator names to their unique Short Codes—compact, token-efficient identifiers (e.g., mapping "13-day Bull Power" to \texttt{DELAY(HIGH,\allowbreak 1)-\allowbreak DELAY(\allowbreak EMA(CLOSE,13),1)} that serve as unambiguous keys in the database schema.
By injecting this enriched context (Name + Short Code) alongside the original strategy text, it prompts an LLM to generate a single executable SQL query. 
The agent then performs an execution-based validation loop to ensure that the statement runs successfully on the PostgreSQL database and returns a non-empty result. 
The finalized SQL string and its corresponding model message are preserved in the shared storage; further engineering details, including the exact SQL prompting pattern and error-handling logic, are described in Appendix~\ref{appendix:AgentFlowAndCode}.


\subsection{Coder}

The Coder serves as the execution endpoint that transforms the retrieved data into final backtesting results. 
Its input consists of three key elements: the user's natural language strategy, the mapped indicator context (names and Short Codes), and a data preview (head and tail rows) of the DataFrame obtained by the Retriever. 
Guided by a system prompt that enforces strict backtesting protocols, the Coder invokes a Python execution tool where the full DataFrame has been pre-injected into the runtime environment. 
The LLM then iteratively writes and debugs the code to calculate the required metrics or execute the strategy logic. 
Full implementation details and the answer extraction procedure are documented in Appendix~\ref{appendix:AgentFlowAndCode}.

\begin{table*}[]
\centering
\caption{This table reports model performance on the synthetic BacktestBench dataset and excludes results from the expert-crafted subset. Overall performance comparison across four task categories. OA denotes the overall performance aggregated across these four categories. Abbreviations: MC: Metrics Calculation, TS: Ticker Selection, PC: Parameter Confirmation, SS: Strategy Selection, ECR: Execution Correctness Rate, EA: Execution Accuracy.} 
\label{tab:overall_performence}
\renewcommand{\arraystretch}{0.75}
\setlength{\tabcolsep}{8pt}
\begin{tabular}{l ccccc cc cccc}
\toprule
\multicolumn{1}{c}{\multirow{2}{*}{\textbf{Model}}} & \multicolumn{5}{c}{\textbf{Back Test Accuracy}} & \multicolumn{2}{c}{\textbf{SQL Generation}} & \multicolumn{4}{c}{\textbf{Indicator Retrieval}} \\ 
\cmidrule(lr){2-6} \cmidrule(lr){7-8} \cmidrule(lr){9-12}
 & OA & MC & TS & PC & SS & ECR & EA & Acc & P & R & F1 \\ 
\midrule

\rowcolor{mycolor}
\multicolumn{12}{c}{\textbf{Open-Source Large Language Models}} \\ 
\midrule

Qwen3 235B & \underline{55.01} & \underline{34.71} & \textbf{82.26} & \underline{75.93} & \textbf{85.04} & \textbf{100.00} & \textbf{97.11} & \underline{76.41} & \underline{93.62} & 94.75 & \underline{94.18} \\
\addlinespace

Qwen3 Next 80B & 44.06 & 26.02 & 67.40 & 65.86 & 68.17 & 99.92 & 94.79 & 73.80 & 92.57 & 94.54 & 93.54 \\
\addlinespace

Qwen3 32B & 47.81 & 27.93 & 75.93 & 70.34 & 72.21 & \textbf{100.00} & 90.72  & 67.41 & 90.44 & 93.99 & 92.18  \\
\addlinespace

Qwen3 30B & 47.03 & 28.67 & 71.25 & 65.86 & 75.06 & 99.97 & 92.86 & 69.73 & 91.78 & 93.25 & 92.51 \\
\addlinespace

Qwen3 14B & 36.97 & 16.96 & 62.04 & 61.57 & 64.61 & 99.95 & 87.59 & 60.92 & 87.56 & 90.50 & 89.01 \\
\addlinespace

Qwen3 8B & 26.36 & 4.93 & 53.92 & 50.56 & 57.48 & 99.97 & 86.89 & 60.61 & 87.33 & 90.66 & 88.96 \\
\addlinespace

Qwen3 4B & 19.63 & 1.77 & 39.06 & 42.54 & 48.22 & 98.85 & 69.94 & 58.34 & 87.16 & 89.84 & 88.48 \\
\addlinespace

Seed OSS 36B & 11.34 & 4.60 & 20.08 & 13.06 & 28.50 & 91.16 & 87.96 & 68.07 & 90.49 & 91.89 & 91.18 \\
\addlinespace

Ministral 3 14B & 26.98 & 7.76 & 47.73 & 50.93 & 58.91 & 97.18 & 72.26 & 59.62 & 86.77 & 90.75 & 88.71 \\
\addlinespace

Ministral 3 8B & 17.26 & 2.83 & 34.11 & 37.31 & 36.34 & 92.23 & 46.35 & 47.99 & 82.20 & 87.69 & 84.86 \\
\addlinespace

Kimi K2 Thinking & 44.97 & 30.02 & 66.57 & 58.02 & 67.46  & 98.20 & 96.06 & \textbf{79.82} &  \textbf{95.09} & \textbf{96.86} & \textbf{95.96} \\
\addlinespace

Kimi Linear 48B & 4.87 & 0.19 & 10.18 & 8.58 & 14.96 & 99.22 & 48.46 & 74.37 & 79.38 & 82.05 & 80.69 \\
\addlinespace

GLM 4.7 & \textbf{56.83} & \textbf{39.13} & 80.33 & \textbf{77.24} & 80.76 & 99.92 & 94.86 & 74.35 & 93.48 & 94.53 & 94.00 \\
\addlinespace

GLM 4.7 Flash & 20.26 & 4.41 & 47.59 & 37.31 & 32.30  & 97.81 & 56.80 & 56.18 & 88.03 & 90.80 & 89.39  \\
\addlinespace

MiniMax M2.1 & 45.91 & 21.98 & \underline{81.71} & 69.22 & 76.72 & 99.56 & 95.02 & 57.09 & 86.65 & 90.76 & 88.66 \\
\addlinespace

GPT OSS 120B & 49.56 & 30.86 & 74.00 & 70.34 & 76.48 & 99.43 & \underline{96.58} & 67.65 & 90.46 & 93.35 & 91.88 \\
\addlinespace

GPT OSS 20B & 33.13 & 17.33 & 51.17 & 49.44 & 62.00 & 94.34 & 90.56 & 67.54 & 90.86 & 92.76 & 91.80 \\
\addlinespace

Mimo V2 Flash & 40.43 & 17.43 & 73.04 & 64.37 & 71.26 & 97.84 & 80.42 & 58.58 & 86.67 & 93.38 & 89.90  \\
\addlinespace

Deepseek V3.2 & 50.70 & 29.18 & 79.78 & 73.51 & \underline{81.47} & 99.97 & 92.86 & 68.98 & 91.01 & \underline{94.88} & 92.90  \\

\midrule

\rowcolor{mycolor}
\multicolumn{12}{c}{\textbf{Closed-Source Large Language Models}} \\ 
\midrule

Gemini 3 Pro & \textbf{67.41} & \textbf{51.67} & \textbf{90.37} & \textbf{82.46} & \textbf{89.07} & 99.64 & \textbf{98.70} & \textbf{87.15} & \textbf{96.08} & 96.07 & \textbf{96.08} \\
\addlinespace

Qwen3 Max & 61.84 & 44.14 & 85.56 & 82.28 & 85.27 & \textbf{100} & 94.73 & 72.42 & 91.75 & 95.82 & 93.74 \\
\addlinespace

Qwen3 Coder Plus & 42.05 & 20.59 & 63.41 & 71.08 & 77.91 & 99.95 & 94.03 & 65.09 & 89.94 & 94.24 & 92.04 \\
\addlinespace

Seed 1.8 & 60.58 & 43.22 & 85.83 & 77.05 & 84.80 & 99.40 & 97.45 & 76.93 & 94.27 & \textbf{96.24} & 95.25 \\
\addlinespace

\bottomrule
\end{tabular}
\end{table*}

\section{Experiments}
\subsection{Experimental Setup.}
\subsubsection{Model Configuration.}

Our experimental benchmark encompasses 23 LLMs, categorized into open-source and closed-source families. 
The open-source cohort spans several major families, including the Qwen3 series (ranging from 4B to the 235B A22B Mixture-of-Experts), the Ministral reasoning series (14B and 8B), the Kimi family (K2-Thinking and Linear 48B), the GLM family (4.7 and Flash), and the GPT-OSS family (120B and 20B), alongside standalone models such as MiniMax-M2.1, Seed OSS 36B, Mimo V2 Flash~\cite{xiao2026mimov2flashtechnicalreport}, and DeepSeek V3.2~\cite{DBLP:journals/corr/abs-2512-02556}. 
The closed-source cohort comprises four leading proprietary models: Gemini 3 Pro, Qwen3 Max, Qwen3 Coder Plus, and Seed 1.8. 
See Appendix~\ref{appendix:Model Deployments} for detailed deployment specifications.
Notably, while most models leverage CoT reasoning, Kimi Linear 48B and Seed OSS 36B operate without CoT. 
To ensure fair comparability and balance generation diversity with stability, we uniformly set the temperature parameter to 0.6 across all inference tasks.

\subsubsection{Evaluation Metrics.}
Regarding evaluation metrics, this study quantifies model performance across three dimensions. First, for the factor retrieval task, we employ Accuracy, Precision, Recall, and F1 Score for comprehensive assessment. Second, for the SQL generation task, in addition to computing the Executable Rate (ECR) of generated SQL statements, we also focus on Execution Accuracy (EA) \cite{DBLP:conf/emnlp/YuZYYWLMLYRZR18}, which is defined as the degree of match between the result set returned by the generated SQL query and the ground-truth result set, ignoring element ordering. Finally, for the four categories of strategy backtesting problems, we report the accuracy of the final execution results produced by the model-generated strategy code, reflecting the end-to-end holistic performance. For metrics calculation problems, a prediction is counted as correct only if the absolute error between the predicted value and the ground-truth value is below \(10^{-3}\); otherwise, it is treated as incorrect.

\subsection{Main Results}

\subsubsection{Dominance of Closed-Source Models and the Catch-up of Open-Source Models}

Table \ref{tab:overall_performence} reveals that the closed-source model Gemini 3 Pro \cite{gemmateam2025gemma3technicalreport} holds a dominant position in overall performance, achieving an Overall Accuracy (OA) of 67.41\%. Notably, in the Metrics Calculation (MC) task, which demands the highest level of logical reasoning, it secures a top score of 51.67\%. In contrast, while the best-performing open-source model GLM 4.7 achieves an OA of 56.83\% and an MC score of 39.13\%, it lags behind Gemini 3 Pro by 10.58 percentage points in OA and 12.54 percentage points in MC. This gap indicates that open-source models continue to face significant bottlenecks in complex financial logic operations.

\subsubsection{Sensitivity Differences to Scaling Laws Across Tasks}

Data analysis based on the Qwen3 series (ranging from 4B to 235B parameters) in Table \ref{tab:overall_performence} uncovers significant differences in sensitivity to model parameter size across different tasks. Logical reasoning tasks are highly sensitive to parameter scale: when the model size is reduced from 235B to 4B, the Metrics Calculation (MC) score plummets from 34.71\% to 1.77\%, resulting in an almost complete loss of computational capability. Conversely, indicator retrieval tasks exhibit lower sensitivity to parameter scale: in the same comparison group, the F1 score for Indicator Retrieval only declined from 94.18\% to 88.48\%, with Qwen3 4B maintaining a relatively high retrieval standard. This suggests that while small-parameter models struggle to complete complex strategy logic calculations, they still possess good potential for basic information retrieval.

\subsubsection{Decoupling of Syntax and Logic Capabilities Due to Lack of CoT}

The evaluation of non-CoT models in Table \ref{tab:overall_performence} highlights a significant decoupling between syntactic proficiency and logical reasoning. Kimi Linear 48B exemplifies this with a high Execution Correctness Rate (ECR) of 99.22\% but a low Execution Accuracy (EA) of 48.46\% and a minimal Metrics Calculation (MC) score of 0.19\%, indicating the generation of executable yet logically flawed code. Similarly, Seed OSS 36B achieves a strong EA of 87.96\% but a low Overall Accuracy (OA) of 11.34\%, demonstrating that successful data retrieval fails to guarantee downstream reasoning success. These results confirm that CoT is essential for integrating code generation with the rigorous logical deduction necessary for quantitative backtesting.

\begin{figure}[t]
 \centering
 \includegraphics[width=\columnwidth]{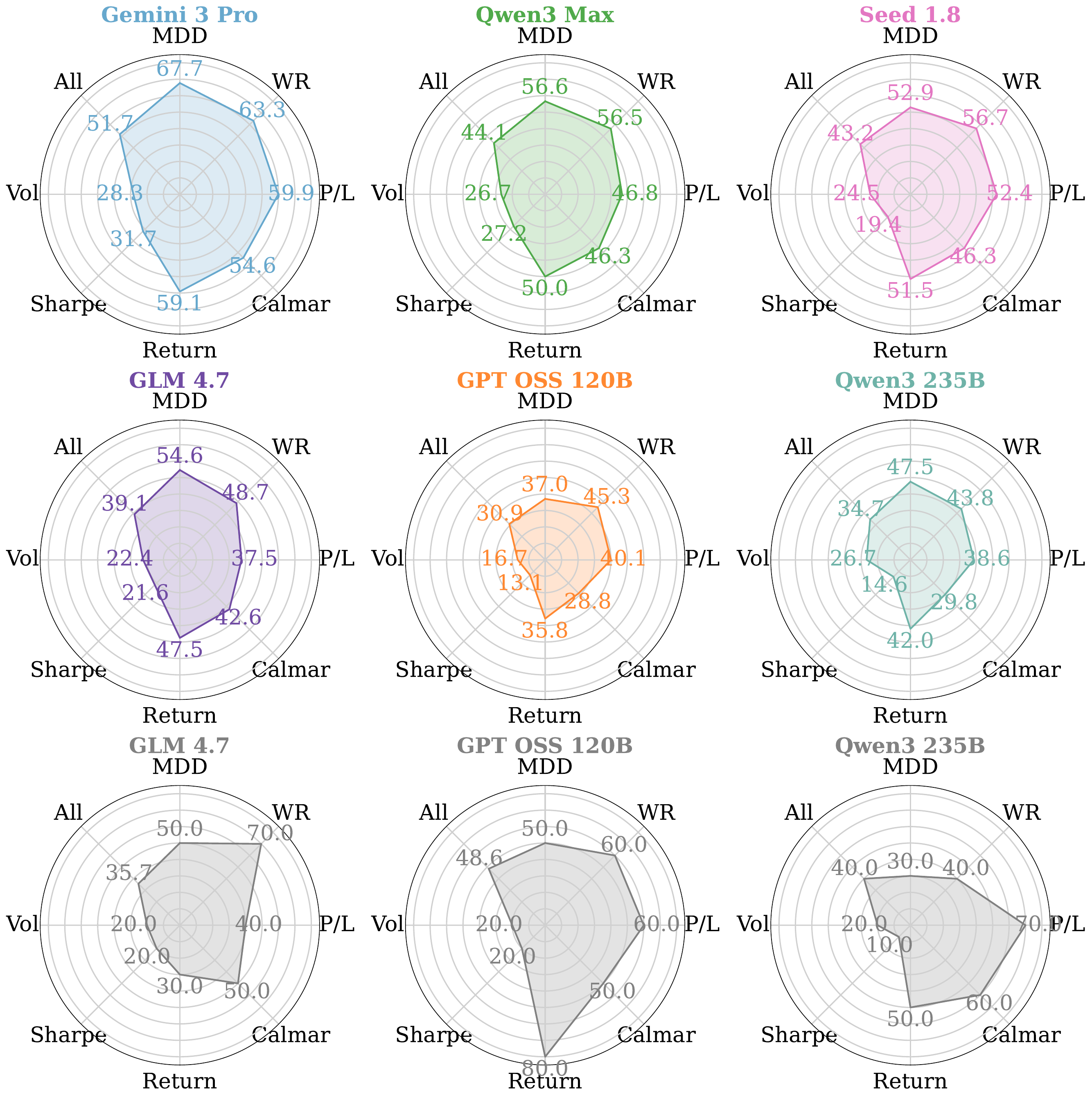}
 \Description{}
 \centering
 \caption{Detailed model performance on the Metrics Calculation task. All denotes the overall performance on the entire Metrics Calculation task. Abbreviations: Sharpe: Annual Sharpe Ratio, Calmar: Calmar Ratio, MDD: Maximum Drawdown, P/L: Profit Loss Ratio, Return: Return Ratio, Vol: Volatility, WR: Win Rate. The first two rows summarize results on the synthetic BacktestBench dataset, while the grey radar chart in the last row reports performance on the expert-crafted evaluation set.}
 \label{fig:All index performance}
\end{figure}

\subsubsection{Accuracy Analysis of Metrics Calculation Tasks}

For metrics calculation problems involving seven types of KPIs, we visualized the performance of the top 3 open-source and top 3 closed-source models (see Figure~\ref{fig:All index performance}).

\paragraph{Complex Mathematical Logic is the Biggest Bottleneck.}
Figure \ref{fig:All index performance} highlights a steep decline in model performance as KPI complexity increases. While logically simpler metrics like Win Rate and Maximum Drawdown see higher accuracy, complex statistical indicators such as Volatility and Sharpe Ratio remain ``disaster zones'' across all models, with significantly lower pass rates. This trend confirms that as tasks escalate from basic arithmetic to intricate statistical operations (e.g., variance, annualization), the reliability of LLM-generated code decays precipitously (see Appendix~\ref{appendix:Case Study} for detailed error analysis).

\paragraph{Closed-Source Models' Moat on ``High-Order Logic''.}
In Figure \ref{fig:All index performance}, Gemini 3 Pro leads with a comprehensive accuracy (All) of 51.67\%, with its core advantage lying in its mastery of high-difficulty indicators. On the most error-prone Sharpe and Vol indicators, Gemini 3 Pro maintains a significant lead (approximately 4.48 and 1.62 percentage points ahead of the second-place Qwen3 Max, respectively). This implies that top-tier closed-source models possess stronger logical rigor when understanding complex financial mathematical formulas and translating them into precise Python/SQL implementations, capable of handling details such as annualization coefficient adjustments and division-by-zero protection.

\paragraph{Open-Source Models: Strong in Extremes, Weak in Statistics.}
GLM 4.7, representing open-source models (All 39.13\%), achieved an accuracy of 54.55\% on MDD (Maximum Drawdown) calculation, a score that even surpasses the closed-source model Seed 1.8 (52.86\%). The core logic of MDD is ``tracking the maximum decline from historical net value peaks,'' which is a typical ``extreme value tracking'' logic. However, on the Vol KPI involving ``distribution statistics'' logic, GLM 4.7 only scored 22.37\%. This contrast illustrates that top open-source models are already perfectly capable of generating procedural calculation code with clear logic but still exhibit high error rates (e.g., formula misuse or library function call errors) when dealing with abstract statistical calculations.

\paragraph{``Logic Collapse'' Zone of Base Models.}
Taking GPT OSS 120B as an example, its scores on Vol and Sharpe KPIs were only 16.71\% and 13.06\%, respectively. This indicates that some models almost completely lose correctness in code implementation when facing complex financial calculations. Such low scores likely stem from the models' inability to correctly understand the conversion relationship between ``annualized volatility'' and ``daily volatility,'' or ignoring data preprocessing (such as logarithmic return transformation) during coding, resulting in calculation results that deviate significantly from standard values.

\paragraph{Synthetic Data Faithfully Mirrors Expert-crafted Subset Difficulty.}
Beyond absolute scores, comparing the synthetic and expert-crafted subset radar plots in Figure \ref{fig:All index performance} reveals that their outer contours are highly aligned: models consistently perform worst on Vol and Sharpe, while achieving much higher accuracy on simpler KPIs. This shape-level consistency indicates that the synthetic BacktestBench data faithfully preserves the intrinsic difficulty hierarchy of real-world backtest metrics, suggesting that both datasets share a similar underlying task distribution rather than reflecting artifacts of our data generation pipeline.

\begin{figure}[t]
 \centering
\includegraphics[width=0.95\columnwidth]{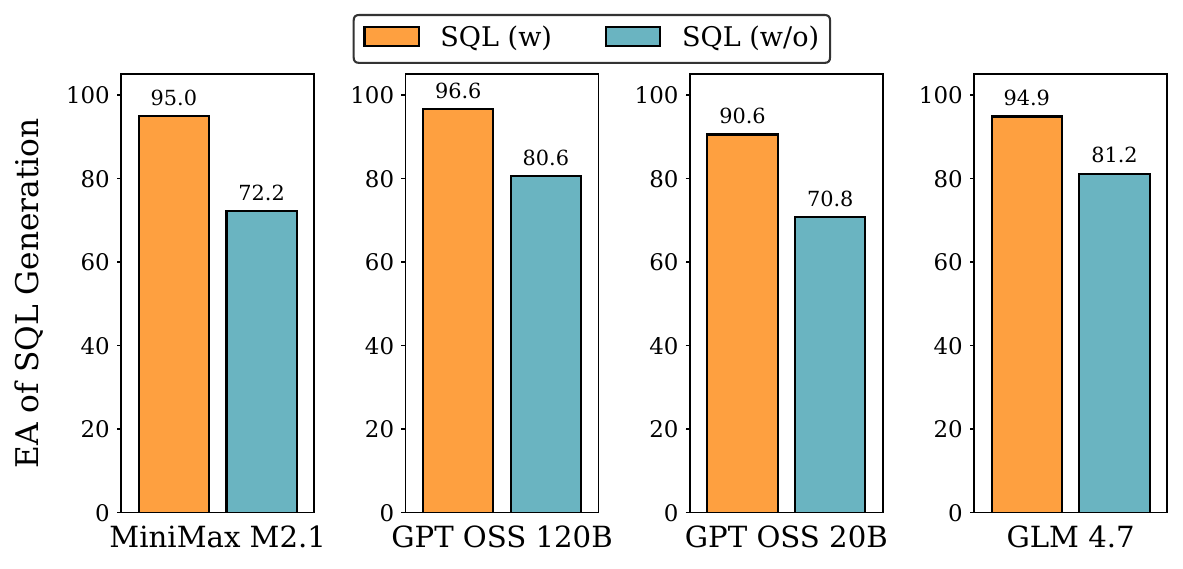}
\Description{A bar chart comparing four models (MiniMax M2.1, GPT OSS 120B, GPT OSS 20B, and GLM 4.7) across two metrics: Back Test Accuracy and Intermediate SQL Generation. For each model, the chart shows comparisons between 'With Factor' (blue bars) and 'Without Factor' (orange bars). The 'With Factor' settings consistently show higher scores than 'Without Factor'.}
 \centering
 \caption{Short Code Ablation.}
 \label{fig:Short_Code_Ablation}
\end{figure}

\begin{figure}[t]
 \centering
\includegraphics[width=\columnwidth]{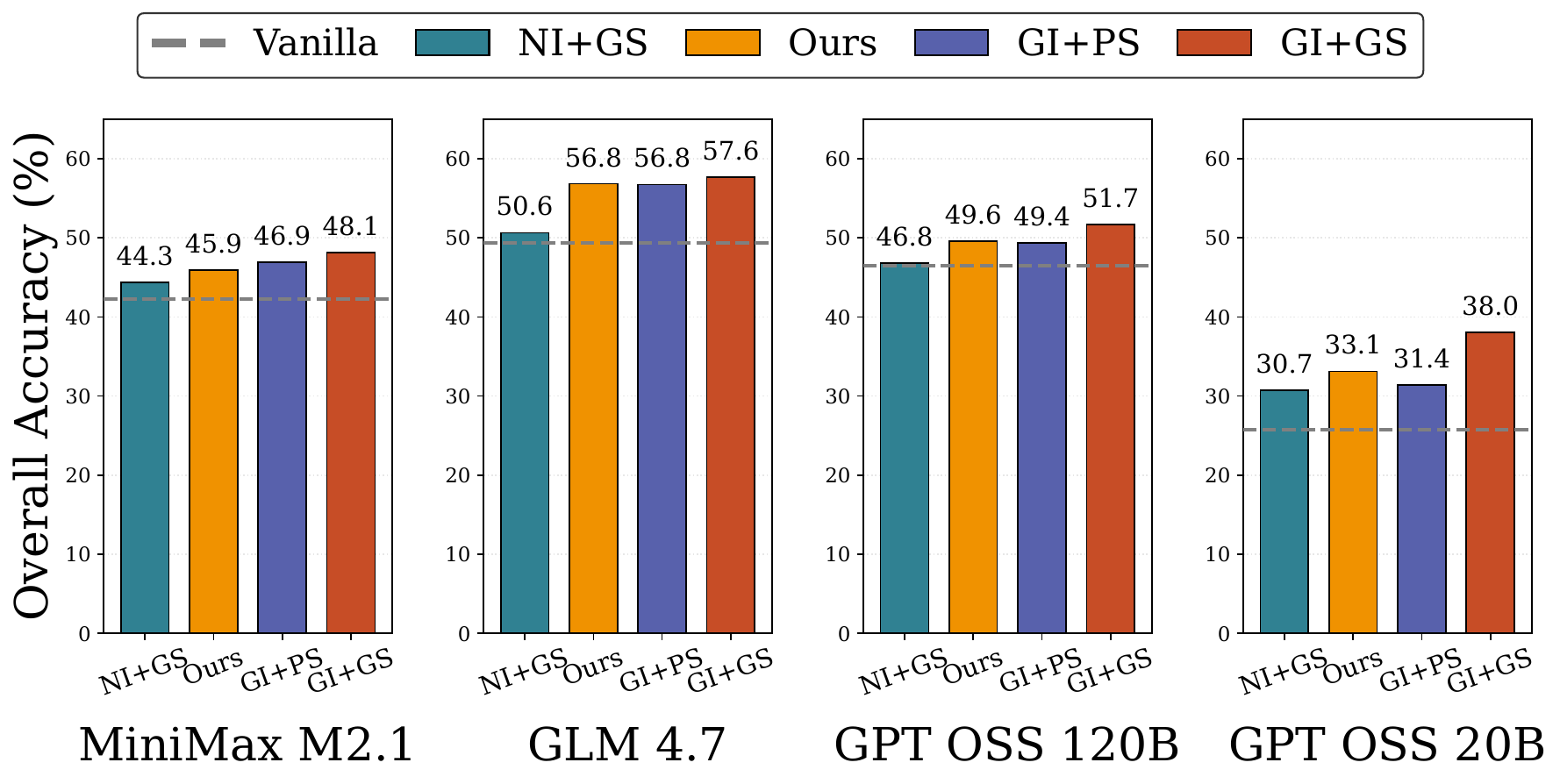}
\Description{}
 \centering
 \caption{Ground Truth Ablation.}
 \label{fig:Ground_Truth_Ablation}
\end{figure}

\subsection{Ablation Study}

\subsubsection{Impact of Short Code on SQL Generation}
We analyze the impact of the Short Code mechanism by comparing SQL generation performance with and without its inclusion. 
As shown in Figure~\ref{fig:Short_Code_Ablation}, the introduction of Short Codes consistently enhances Execution Accuracy (EA) across all evaluated models. 
Specifically, MiniMax M2.1 and GPT OSS 20B exhibit the most pronounced performance jumps, with EA improvements exceeding 20 percentage points, while robust performers like GPT OSS 120B and GLM 4.7 also achieve clear gains.
These results confirm that Short Codes act as critical semantic anchors, effectively guiding the models to map natural language intents to precise database schema elements.

\subsubsection{Independent Impact of Short Code and SQL Quality}

To deeply investigate the independent effects of Short Code and SQL generation quality on strategy backtesting performance, we designed five experimental configurations for comparative analysis. \textbf{Vanilla} serves as the baseline, where the LLM generates SQL and backtesting code directly from the user query without any Short Code information. \textbf{Ours} represents the complete implementation of AutoBacktest, including retrieval-augmented Short Code context. Additionally, we introduce three variants: \textbf{GI+PS} (Gold Indicator + Predicted SQL) uses the ground-truth Short Code (Gold Indicator) as context but lets the LLM predict the SQL; \textbf{NI+GS} (No Indicator + Gold SQL) uses no Short Code but provides the ground-truth SQL (Gold SQL); and \textbf{GI+GS} (Gold Indicator + Gold SQL) provides both ground-truth Short Code and Gold SQL, serving as the theoretical performance upper bound.

The experimental results in Figure~\ref{fig:Ground_Truth_Ablation} reveal the decisive role of Short Code in the strategy generation pipeline. Overall, all configurations incorporating Short Code information (\textbf{Ours}, \textbf{GI+PS}, \textbf{GI+GS}) significantly outperform those without it (\textbf{Vanilla}, \textbf{NI+GS}). Remarkably, although the \textbf{NI+GS} method uses perfect Gold SQL, its performance, while better than the Vanilla baseline, still lags substantially behind the three groups containing Short Code. This phenomenon strongly demonstrates that including Short Code in the prompt significantly enhances the LLM's ability to understand and apply financial indicators, yielding performance gains far exceeding those from optimizing SQL statements alone. Furthermore, comparing \textbf{Ours} with \textbf{GI+PS} reveals minimal performance differences, with \textbf{Ours} even slightly outperforming in some GPT OSS models (likely due to random experimental error). This further indicates that given accurate Short Code context, the LLM can generate correct final backtesting code through contextual understanding even if the generated SQL has minor imperfections, highlighting the robustness of the AutoBacktest framework against SQL errors.

\section{Related Work}

\subsection{LLM-based Strategy Backtesting}


Recent works have explored LLMs for quantitative tasks, yet gaps remain in backtesting rigor. Automate Strategy Finding \cite{DBLP:journals/corr/abs-2409-06289} and QuantAgent \cite{DBLP:journals/corr/abs-2402-03755} focus on factor mining and self-improving trading agents, prioritizing profitability over the standardization and reproducibility of the underlying code execution. Similarly, FinMem \cite{DBLP:conf/aaaiss/YuLCJLZLSK24} addresses long-term memory for trading decisions but overlooks the challenges of complex data retrieval and metrics calculation. While AutoPrep \cite{DBLP:journals/pvldb/FanFTCLD25} automates general tabular data preprocessing, it lacks the specialized temporal logic required to prevent look-ahead bias in financial backtesting.

\subsection{Datasets for Strategy Backtesting}


Existing benchmarks primarily target Reinforcement Learning or specific sub-tasks rather than the full backtesting workflow. TradeMaster \cite{NEURIPS2023_b8f6f7f2} and FinRL-Meta \cite{DBLP:conf/nips/LiuXRGYZWWG22} provide extensive market environments for training RL agents but do not evaluate the generation of interpretable strategy logic. FNSPID \cite{DBLP:conf/kdd/DongFP24} aligns news with market data but serves mainly as an information retrieval resource. 
While QuantEval \cite{kang2026quantevalbenchmarkfinancialquantitative} assesses strategy coding, it remains limited to a small scale of 60 problems. Similarly, StockBench \cite{DBLP:journals/corr/abs-2510-02209} prioritizes the profitability of agentic trading decisions, and Market-Bench \cite{DBLP:journals/corr/abs-2512-12264} evaluates the implementation accuracy of introductory strategies. In contrast, BacktestBench pioneers a large-scale evaluation of the end-to-end quantitative research workflow. It spans natural language understanding, SQL retrieval, and rigorous backtest execution, comprehensively assessing the translation of complex natural language into executable backtesting logic.

\section{Conclusion}

In this paper, we introduce BacktestBench, a pioneering benchmark dedicated to automated quantitative strategy backtesting, a domain characterized by complex temporal logic and rigorous precision requirements. To support this initiative, we construct a large-scale dataset with 18,246 high-quality QA pairs derived from real-world market records, covering four core decision-making tasks: metrics calculation, ticker selection, strategy selection, and parameter confirmation. Complementing this resource, we propose AutoBacktest, a multi-agent collaboration framework that mimics the professional workflow of quantitative researchers to achieve end-to-end automation from natural language strategy descriptions to executable backtesting code.
We systematically evaluate 23 mainstream LLMs on BacktestBench and conduct comprehensive ablation studies to identify key performance drivers in this domain. By establishing a standardized evaluation protocol and a rigorous data resource, this work bridges the gap between general-purpose code generation and quantitative investment research, laying a solid foundation for future advancements in intelligent financial decision-making.

\begin{acks}

This work is supported in part by the National Natural Science Foundation
of China (NSFC) under Grant 62272050 and the grant of Beijing Normal-
Hong Kong Baptist University sponsored
by Guangdong Provincial Department of Education;  in part by Zhuhai Science-Tech Innovation Bureau under
Grant No. 2320004002772 and the Interdisciplinary Intelligence
Super Computer Center of Beijing Normal University (Zhuhai).

\end{acks}

\bibliographystyle{ACM-Reference-Format}
\bibliography{references}

\appendix

\section{Factors}
\label{appendix:Factors}

This section provides a comprehensive reference for all factors and KPIs used throughout BacktestBench. Table~\ref{tab:all_factors_used} presents the complete inventory of factors employed in the benchmark, while Tables~\ref{tab:all_KPI_used} and~\ref{tab:KPIs with Short Code.} detail the correspondence between full names and short codes for factors and KPIs, respectively.

\begin{figure*}[t]
 \centering
\includegraphics[width=0.85\textwidth]{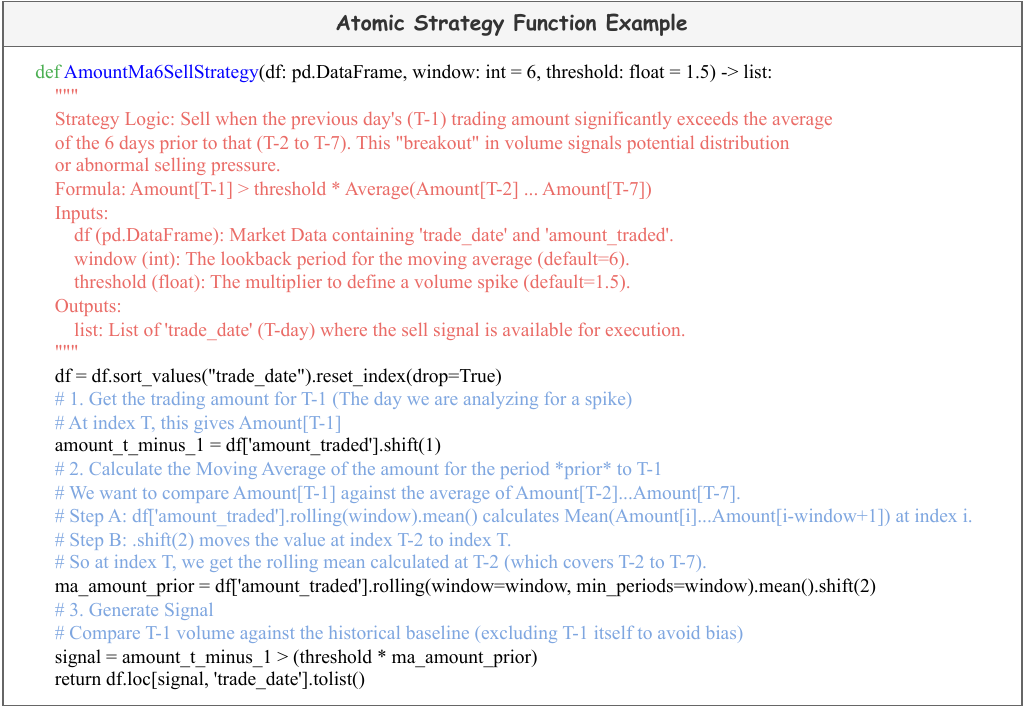}
\Description{Atomic Strategy Function Example}
 \centering
 \caption{Atomic Strategy Function Example.}
 \label{fig:Atomic Strategy Function Example}
\end{figure*}

We additionally provide an atomic strategy function example in Figure~\ref{fig:Atomic Strategy Function Example}. The example \texttt{AmountMa6SellStrategy} triggers a sell signal when the trading volume on day \( T-1 \) significantly exceeds the average volume of the preceding \( 6 \) days (from \( T-2 \) to \( T-7 \)). Here, the \textit{threshold} parameter serves as a multiplier to define what constitutes a significant volume spike; a signal is generated only if the volume at \( T-1 \) is greater than the historical average multiplied by this threshold. This design strictly avoids look-ahead bias by comparing the observed volume at \( T-1 \) against a baseline computed solely from data prior to \( T-1 \), ensuring the decision at day \( T \) relies only on historically available information.

It is important to note that factors appearing similar but differing in time windows (e.g., 5-day vs. 20-day moving averages) are treated as distinct factors in our benchmark. This distinction is crucial because these variations capture fundamentally different market dynamics: shorter windows reflect short-term momentum and volatility patterns, while longer windows indicate long-term trends and structural movements. Consequently, each time-window variant represents a unique factor with distinct predictive characteristics and cannot be considered equivalent or interchangeable.

The factor and KPI naming system serves a critical role in the multi-stage AutoBacktest pipeline. During the Summarizer stage, the system retrieves and identifies indicators (include factors and KPIs) using their full names (e.g., ``5-day Moving Average" or ``Annual Sharpe Ratio"). Subsequently, in the Retriever and Coder stages, both the full names and their corresponding short codes are injected into the prompt context. This dual representation enables the LLM to understand the semantic meaning through full names while generating concise, standardized code using short codes, thereby ensuring consistency and reducing ambiguity in the automated backtesting process.

\section{Backtesting Protocol and Standardized Trading Rules}
\label{appendix:BacktestingProtocol}
Several financial KPIs are sensitive to implementation details and can be mis-specified in non-standardized settings. Moreover, LLMs are prone to hallucinations when reasoning about these calculations. Therefore, we define a standardized backtesting and trading protocol that enforces deterministic rules and unambiguous computational logic, ensuring that the backtest results are unique and reproducible.

\paragraph{Execution Mechanism.}
The backtesting framework adopts a long-only trading scheme with strictly alternating buy and sell operations, and it explicitly disallows intraday round trips (T+0).
All trades execute at observable and well-defined prices: buy orders fill at the opening price (Buy at Open) and sell orders fill at the closing price (Sell at Close), which removes ambiguity about intraday price paths.
If a position remains open on the final day of the backtest window, the framework forcibly liquidates it at the corresponding closing price.

\paragraph{Allocation and Constraints.}
The simulation uses a fixed-capital setup with initial wealth denoted by \(\text{Cash}_0\), and it does not allow leverage.
On each buy date \(t\), the trade size \(Q_t\) is computed dynamically as
\[
Q_t = \left\lfloor \frac{\text{Cash}_t}{\text{Price}_{\text{open}, t}} \right\rfloor,
\]
subject to a minimum order lot constraint \(Q_t \in \mathbb{Z}_{\ge 100}\), which corresponds to trading in round lots of 100 shares.
On sell dates, the framework enforces full liquidation of the current position.
To isolate the impact of trading logic, the baseline experiments set transaction costs and taxes to zero.

\paragraph{Performance Evaluation.}
The framework adopts a daily mark-to-market convention when computing returns.
The portfolio value \(\text{PV}_t\) at date \(t\) is the sum of the market value of all open positions marked at the closing price and the remaining cash balance.
The daily return is computed as
\[
r_t = \frac{\text{PV}_t - \text{PV}_{t-1}}{\text{PV}_{t-1}},
\]
with \(r_t = 0\) on days without an open position.
Annualized return and annualized volatility follow the convention of 252 trading days per year, and the daily risk-free rate is fixed to 0.0001 when evaluating the Annual Sharpe Ratio.
The Calmar Ratio is computed as the ratio of annualized return to maximum drawdown, which emphasizes downside-risk-adjusted performance.

\paragraph{Data Integrity Assumption.}
The backtest assumes that the input price time series is complete and strictly ordered by trading date.
The current framework does not explicitly handle missing trading days, irregular calendars, or out-of-order records, and we leave these data quality issues for future extensions.

\section{Datasets}
\label{appendix:Dataset}

\subsection{Database Schema}
\label{appendix:Database}

Table~\ref{tab:db_schema} presents the schema of the \texttt{stocks} PostgreSQL~\cite{postGRE} database, detailing the 15 constituent columns along with their data types and descriptions. This unified schema is applied consistently across the following three exchange tables:
\begin{itemize}
    \item \texttt{beijing\_stock\_exchange}
    \item \texttt{shenzhen\_stock\_exchange}
    \item \texttt{shanghai\_stock\_exchange}
\end{itemize}

\begin{table}[t]
  \centering
  \caption{Schema of the \texttt{stocks} database.}
  \label{tab:db_schema}
  \begin{tabular*}{\columnwidth}{@{\extracolsep{\fill}} l l}
    \toprule
    \textbf{Column} & \textbf{Type} \\
    \midrule
    \texttt{name}                     & TEXT    \\
    \texttt{stock\_code}              & TEXT    \\
    \texttt{trade\_date}              & DATE    \\
    \texttt{opening\_price}           & NUMERIC \\
    \texttt{closing\_price}           & NUMERIC \\
    \texttt{highest\_price}           & NUMERIC \\
    \texttt{lowest\_price}            & NUMERIC \\
    \texttt{volume\_traded}           & NUMERIC \\
    \texttt{amount\_traded}           & NUMERIC \\
    \texttt{percentage\_change}       & NUMERIC \\
    \bottomrule
  \end{tabular*}
\end{table}

\begin{table}[t]
\centering
\caption{Overall dataset statistics for BacktestBench across four core Quantitative Investing task families: metrics calculation, ticker selection, parameter confirmation, and strategy\_selection.}
\label{tab:dataset-statistics}
\begin{tabular*}{\columnwidth}{@{\extracolsep{\fill}} l r}
\toprule
\textbf{Statistics} & \textbf{Num} \\
\midrule
\# Total Questions              & 18{,}246 \\
\# Train                        & 10{,}215 \\
\# Validation                   & 4{,}195 \\
\# Test                         & 3{,}836 \\
\# Metrics Calculation          & 10{,}244 \\
\# Ticker Selection             & 3{,}455 \\
\# Parameter Confirmation       & 2{,}548 \\
\# Strategy Selection          & 1{,}999 \\
\# Avg. Length per Strategy     & 141.7 \\
\bottomrule
\end{tabular*}
\end{table}

\subsection{Strategy Data Generation Details}
\label{appendix:DataGenDetails}

This subsection details the construction pipeline for the four task families introduced in Section~3.
Unless otherwise specified, all strategy instances are generated from the signal pool and KPI library described in Section~\ref{appendix:Factors} under the standardized backtesting protocol.

\paragraph{Foundational Framework and Metrics Calculation Tasks.}
For metrics calculation tasks, the generation pipeline consists of three stages: signal pool sampling, logic composition, and backtest-based validation.
First, the system samples a single stock from the three Chinese exchanges according to predefined weights and extracts a contiguous historical window of at least 252 trading days, which is standardized into a DataFrame containing OHLCV fields.
Next, using code introspection, the framework independently samples up to four buy-side and four sell-side atomic factors, fuses their signals via set intersection (trades occur only when all selected factors fire on the same date), and randomly chooses one KPI as the optimization objective, assembling them into an executable Python strategy script.
Finally, the script is run in an isolated interpreter, the KPI value is recorded as the ground-truth answer, and low-quality samples with sparse trades or anomalous outputs (e.g., \texttt{NaN} or infinite values) are filtered out, yielding a total of \textbf{17,082} metrics calculation programs.

\paragraph{Ticker Selection Tasks.}
Ticker selection tasks extend the above framework to a multi-asset setting.
Instead of a single underlying, the system samples a candidate set of stocks, applies the same randomly generated strategy and KPI objective to each asset in parallel, and runs backtests under identical experimental conditions.
According to the economic meaning of the KPI, the framework adopts either a maximization principle (for reward-oriented metrics such as return or Sharpe Ratio) or a minimization principle (for downside or risk metrics such as Maximum Drawdown or Volatility) to label the best-performing ticker as the ground truth, resulting in \textbf{6,098} ticker selection programs. 

\paragraph{Parameter Confirmation Tasks.}
Parameter confirmation tasks emulate controlled hyperparameter tuning within a fixed strategy and environment.
The system first constructs a baseline strategy with a deterministic factor combination and KPI, then selects one atomic factor (e.g., a moving-average breakout rule) as the optimization target and samples a discrete set of candidate parameter values from a predefined space.
For each candidate, a strategy variant is formed by substituting the target parameter while keeping all other logic unchanged; all variants are backtested on the same price path and evaluated under the same KPI-based selection rule, and the parameter yielding the best KPI is recorded as the label, producing \textbf{4,709} parameter confirmation programs.

\paragraph{Strategy Selection Tasks.}
Strategy selection tasks capture the decision problem of choosing among heterogeneous strategy logics under a shared market environment. 
The framework samples a small set of candidate strategies (e.g., a binary contest $\{A,B\}$ or ternary contest $\{A,B,C\}$), independently generates buy and sell factor combinations for each candidate, and constrains all of them to trade the same underlying over the same historical window with identical initial capital and KPI.
Dynamic code generation then embeds all candidate strategies into a single script and runs parallel backtests; by comparing KPI values under the appropriate maximization or minimization criterion, the system identifies the best-performing logic (such as ``B'') as the ground-truth label, yielding \textbf{5,114} strategy\_selection programs.

\paragraph{SQL Query Construction and Annotation.}
For every strategy instance, the framework automatically constructs and annotates an SQL statement that exactly reproduces the underlying data slice used in backtesting.
All queries follow a standard \texttt{SELECT\allowbreak–FROM\allowbreak–WHERE} pattern: during sampling, the chosen exchange, ticker, and date interval are mapped respectively to the \texttt{FROM} clause, ticker constraint, and time filter (e.g., \texttt{FROM shenzhen\_stock\_exchange WHERE name IN (`Ping An Bank') AND trade\_date BETWEEN `2019-01-15' AND `2020-03-20'}).
After strategy code generation, the system parses the code to extract the actually used columns (such as \texttt{opening\_price}, \texttt{closing\_price}, \texttt{volume\_traded}) and constructs a minimal \texttt{SELECT} clause containing only required fields, thereby avoiding redundant data loading.
The finalized SQL string is stored together with the strategy code, backtest interval, and ticker list in a JSON file, ensuring a one-to-one correspondence between code, data, and environment that facilitates exact reproduction and error tracing.

\subsection{Natural Language Generation and Evaluation Details}
\label{appendix:NL_details}

This subsection provides additional implementation details for the natural-language description generation and quality assessment pipeline introduced in Section~3.

\paragraph{Prompt Design and Generation Setup.}
We employ five specific on-premise models for generation: Kimi-K2-Thinking-BF16, MiniMax-M2.1-BF16, GLM-4.7, GPT-OSS-120B-BF16, and Qwen3-235B-A22B-Thinking-2507. All five models support CoT reasoning and are configured as instruction-following LLMs. The prompts enforce an instructional tone that mimics realistic backtesting requests, explicitly separate strategy-specific logic from system-level defaults (such as capital allocation and execution microstructure), and require precise temporal expressions (e.g., ``based on $T-1$ data'') to avoid look-ahead bias.

\paragraph{Evaluation Criteria and Reporting.}
During evaluation, each of the four evaluator models produces a structured JSON report for every code–text pair, containing binary decisions for code fidelity and strategy validity, together with brief error diagnoses (e.g., ``parameter mismatch'', ``future information usage'', or ``missing exit rule'').
Only descriptions that receive positive judgments from all evaluators are retained, and the remaining reports are used to categorize and analyze common failure modes.

\paragraph{Scale and Resource Usage.}
The full pipeline processes 33,003 strategy programs and issues 825,075 LLM calls across the five models deployed on 72 NVIDIA A800-SXM4-80GB GPUs. Under the strict acceptance rule, the final corpus contains 10,244 metrics calculation tasks, 3,455 ticker selection tasks, 2,548 parameter confirmation tasks, and 1,999 strategy selection tasks, totaling 18,246 samples partitioned into 10,215 training, 4,195 validation, and 3,836 test instances. Additional dataset statistics are presented in Table \ref{tab:dataset-statistics}.

\section{Agent Data Flow and Interaction}
\label{appendix:AgentFlowAndCode}

In the AutoBacktest framework, each strategy instance is represented as a shared record initialized with a natural-language description stored in the \texttt{strategy} field. The Summarizer, Retriever, and Coder agents operate sequentially on this record, progressively transforming the raw text into a fully specified backtesting sample containing factor definitions, executable data queries, and quantitative results.

The Summarizer is the semantic entry point of the system. Its input consists of the fields  strategy, and optionally the global definition of the Financial Indicator vocabulary (which explicitly encompasses both the 43 trading factors and 7 performance KPIs). Based on this information, the Summarizer analyzes the strategy description, extracts the mentioned indicators, normalizes informal wording into canonical Financial Indicator names, and produces a structured list of predicted indicators together with any auxiliary annotations needed for later retrieval. This list is written back into the shared record as fields such as \texttt{predict\_indicators} and a refined version \texttt{predict\_indicators(BM25)}, which serve as the semantic bridge from free-form language to the financial indicator library.

The Retriever builds on the enriched record produced by the Summarizer. Its input includes the original strategy text \texttt{strategy}, the predicted indicator list \texttt{predict\_indicators(BM25)}, and a separate financial indicator dictionary that maps each financial indicator name to its implementation-level identifiers, for example short codes or database column names. Using these inputs, the Retriever constructs a compact textual context that pairs financial indicators with their short codes and then generates a single SQL statement that specifies which ticker, time interval, and raw fields should be fetched from the historical database for this strategy. Before the statement is accepted, the Retriever executes it against the database to check for syntax correctness and non-empty results; if necessary, it iteratively revises the query based on error messages or empty outputs. Once a valid query is obtained, the final SQL string is stored in the shared record as \texttt{predict\_SQL}.

The Coder is the execution endpoint that turns the verified data-access specification into a concrete backtesting result. Its input is the fully enriched record after the Retriever stage, including the strategy text \texttt{strategy}, the strategy type \texttt{strategy\_type} (which determines whether the task is metrics calculation, ticker selection, parameter confirmation, or strategy selection), the indicator context \texttt{predict\_indicators(BM25)} together with the indicator dictionary, and the validated SQL statement \texttt{predict\_SQL}. The Coder first executes \texttt{predict\_SQL} to obtain a time series of market data corresponding to the requested ticker and period; this data is treated as the backtesting environment, and an empty result triggers an early termination for that record with an error flag. If a non-empty data slice is obtained, the Coder constructs a task-specific prompt that combines a preview of the retrieved table, the normalized indicator information, and the strategy description, and then uses a tool-augmented language model to write and execute backtesting code within this environment. The final model message is parsed to extract a standardized answer, such as a KPI value, an index of the best-performing ticker, a selected parameter, or a chosen strategy. The final model message is parsed to extract a standardized answer, such as a KPI value, an index of the best-performing ticker, a selected parameter, or a chosen strategy. The extracted scalar answer is appended to the shared record as the \texttt{pred\_answer} field.

\section{Experiments}
\label{appendix:Experiments}

\begin{table*}[htbp]
    \centering
    \caption{Performance Comparison of the Summarizer Module With and Without BM25 Retrieval. This table evaluates the effectiveness of incorporating the BM25 algorithm into the Summarizer stage across various LLMs. ``w/o" denotes the baseline performance without BM25, while ``w" indicates performance with BM25 integration. The metrics include Accuracy (Acc), Precision (P), Recall (R), and F1 Score.}
    \label{tab:bm25_ablation}
    \setlength{\tabcolsep}{3.5mm}
    \begin{tabular}{lccccc}
    \toprule
    \textbf{Model} & \textbf{BM25 Status} & \textbf{Acc (\%)} & \textbf{P (\%)} & \textbf{R (\%)} & \textbf{F1 (\%)} \\ 
    \midrule
    \multirow{2}{*}{DeepSeek V3.2} & w/o & 17.65 & 64.35 & 69.71 & 66.92 \\
     & w & \textbf{68.98} & \textbf{91.01} & \textbf{94.88} & \textbf{92.90} \\ 
    \midrule
    \multirow{2}{*}{Qwen3 4B} & w/o & 13.03 & 61.24 & 63.06 & 59.52 \\
     & w & \textbf{58.34} & \textbf{87.16} & \textbf{89.84} & \textbf{88.48} \\ 
    \midrule
    \multirow{2}{*}{Qwen3 8B} & w/o & 15.64 & 60.98 & 65.24 & 63.04 \\
     & w & \textbf{60.61} & \textbf{87.33} & \textbf{90.66} & \textbf{88.96} \\ 
    \midrule
    \multirow{2}{*}{Qwen3 14B} & w/o & 11.47 & 57.59 & 59.52 & 55.79 \\
     & w & \textbf{60.92} & \textbf{87.56} & \textbf{90.50} & \textbf{89.01} \\ 
    \midrule
    \multirow{2}{*}{Qwen3 30B} & w/o & 20.36 & 68.10 & 70.04 & 69.06 \\
     & w & \textbf{69.73} & \textbf{91.78} & \textbf{93.25} & \textbf{92.51} \\ 
    \midrule
    \multirow{2}{*}{Qwen3 Next 80B} & w/o & 27.42 & 73.88 & 76.39 & 75.12 \\
     & w & \textbf{73.80} & \textbf{92.57} & \textbf{94.54} & \textbf{93.54} \\ 
    \midrule
    \multirow{2}{*}{Qwen3 235B} & w/o & 30.42 & 75.55 & 77.19 & 76.36 \\
     & w & \textbf{76.41} & \textbf{93.62} & \textbf{94.75} & \textbf{94.18} \\ 
    \midrule
    \multirow{2}{*}{GLM 4.7} & w/o & 24.35 & 72.06 & 73.78 & 72.91 \\
     & w & \textbf{74.35} & \textbf{93.48} & \textbf{94.53} & \textbf{94.00} \\ 
    \midrule
    \multirow{2}{*}{MiniMax M2.1} & w/o & 14.21 & 60.19 & 65.33 & 62.65 \\
     & w & \textbf{57.09} & \textbf{86.65} & \textbf{90.76} & \textbf{88.66} \\ 
    \bottomrule
    \end{tabular}
\end{table*}

\subsection{Impact of BM25 in Summarizer}
\label{appendix:Summarizer Disscution}

To evaluate the necessity of the retrieval mechanism in our framework, we conduct an ablation study on the Summarizer module. Specifically, we compare the performance of various LLMs with and without the integration of the BM25 algorithm. In the configuration without BM25 (``w/o"), the Summarizer relies solely on the LLM's internal knowledge to identify relevant intents and slots. Conversely, the configuration with BM25 (``w") augments the prompt with retrieved context related to indicator definitions and project metadata.

Table~\ref{tab:bm25_ablation} presents the comparative results. The integration of BM25 yields a significant performance improvement across all tested models. For instance, DeepSeek V3.2 exhibits a substantial increase in Accuracy from 17.65\% to 68.98\% and an F1 score improvement from 66.92\% to 92.90\%. Even larger models, such as Qwen3 235B, benefit markedly, with Accuracy rising from 30.42\% to 76.41\%. These results demonstrate that while LLMs possess strong reasoning capabilities, the precise identification of domain-specific indicators and intents in quantitative investing requires external knowledge retrieval. The BM25 algorithm effectively bridges this gap by providing accurate context, thereby significantly enhancing the Summarizer's ability to map natural language queries to correct structured representations.

\subsection{Case Study}
\label{appendix:Case Study}

To diagnose the root causes of performance degradation in \textit{metrics calculation} tasks—specifically Volatility and Sharpe Ratio—we construct a targeted analysis framework for the \textit{backtesting} process. In our experiments with Gemini-3-Pro, we filter for samples with accurate SQL queries and successful \textit{indicator} retrieval, isolating instances where Python code execution produces unexpected deviations. Our analysis identifies two primary mechanisms driving these failures.

\subsubsection{Case 1: Composite Attribution Analysis of Volatility Calculation Failure}
This case examines the failure mechanism in volatility calculation tasks, where three systemic logical defects in the generated code cumulatively distort \textit{backtesting} results. 

The primary error occurs during performance attribution, where the code fails to properly initialize the return series. By neglecting to insert the initial capital as a ``Day 0'' baseline, the first-day return is calculated as invalid and subsequently discarded. This left-side truncation compromises the integrity of the time series, forcing the volatility calculation to rely on an incomplete sample set and introducing significant numerical errors. 

Furthermore, regarding \textit{indicator} construction, the model misinterprets the mathematical definition of ``Mean Absolute Deviation (MAD)'' within the Commodity Channel Index (CCI). Instead of the standard ``average absolute deviation within a rolling window,'' the code implements a ``simple rolling average of absolute price deviations.'' This algorithmic discrepancy causes buy signal triggers to diverge from the intended strategy. 

Additionally, the trading logic lacks necessary boundary constraints for the \textit{backtesting} cycle. Without a filter to prohibit position opening on the final day, the system executes non-compliant intraday bidirectional trading (opening at market start and forced liquidation at close). This artificially inflates transaction costs and skews the final net value, rendering the volatility calculation ineffective.

\subsubsection{Case 2: Systematic Bias in Sharpe Ratio Assessment}
This case investigates the failure mechanism affecting the Sharpe Ratio, a metric whose accuracy relies heavily on the completeness of the return series and strict adherence to trading logic. We observe that logical defects in two key dimensions cause severe distortion in the assessment.

A critical flaw appears in the construction of the net value sequence, where the code omits the ``zero-point anchoring'' operation. Failing to explicitly insert the initial capital at the head of the sequence leads to the exclusion of the first-day return during difference calculations. This reduction in sample size affects both the numerator (annualized excess return) and the denominator (return volatility), resulting in a Sharpe Ratio derived from a fragmented time window and destroying statistical validity.

Simultaneously, the code exhibits insufficient boundary control at the end of the \textit{backtesting} period. By permitting non-compliant ``open and close on the same day'' operations, the system incurs invalid transaction costs that are directly borne by the net value sequence. This artificially depresses the final return performance, depriving the generated Sharpe Ratio of its reference value as a performance benchmark.

\section{Impletation Details}
\label{appendix:Impletation Details}

\subsection{LLM Deployments}
\label{appendix:Model Deployments}

This section details the LLM configurations employed across the different stages of our research. The deployment is divided into two distinct phases: the data construction phase and the experimental evaluation phase.

During the data construction phase, we utilized a high-perform-\allowbreak ance computing cluster equipped with 72 NVIDIA A800-SXM4-80GB GPUs to deploy five state-of-the-art open-source CoT models. These models were selected for their superior reasoning capabilities and include Kimi-K2-Thinking-BF16, MiniMax-M2.1-BF16, GLM-4.7, GPT-OSS-120B-BF16, and Qwen3-235B-A22B-Thinking-2507. Detailed specifications for these models are provided in Table~\ref{tab:LLM_deployment}. To optimize inference efficiency, we leveraged SGLang and vLLM as our deployment frameworks. A unified temperature parameter of 0.6 was applied across all models to maintain consistency in generation diversity and stability.

In the experimental phase, we expanded our infrastructure to 100 NVIDIA A800-SXM4-80GB GPUs to locally host 18 models for comprehensive benchmarking. To complement these local deployments, we accessed several proprietary models via external APIs. Specifically, Mimo V2 Flash and Gemini 3 Pro were accessed through OpenRouter, incurring a total cost of \$756. For DeepSeek V3.2, Qwen3 Max, and Qwen3 Coder Plus, we utilized their official compatible-mode APIs available at \url{https://dashscope.aliyuncs.com/compatible-mode/v1}. Additionally, Seed 1.8 was accessed via the Volcengine API at \url{https://ark.cn-beijing.volces.com/api/v3}. To ensure a fair comparison between local and API-based models, we strictly maintained the temperature setting at 0.6 for all inferences throughout the experiment.

\subsection{Agent Impletation}
\label{appendix:Agent Impletation}

We implement the AutoBacktest agent using the LangGraph framework, employing a ReAct (Reasoning and Acting) architecture to orchestrate the backtesting workflow. The agent interacts with a PostgreSQL database to retrieve financial data, which is converted into a Pandas DataFrame. We equip the agent with a Python REPL tool (`create\_python\_repl\_tool'), enabling it to execute Python code directly on the dataframe to perform complex calculations and logic verification.

To ensure robustness and standardized outputs, the implementation features several key components:
\begin{enumerate}
    \item \textbf{Dynamic Context Construction}: The system dynamically generates prompts by injecting the specific trading strategy, indicator definitions retrieved via the BM25 Summarizer, and the dataframe schema. This ensures the LLM possesses all necessary context for code generation.
    \item \textbf{Structured Output Parsing}: We define specific Pydantic models (e.g., \texttt{ClassA\_FinalAnswerFormat}) for different task types, such as metric calculation or ticker selection. This enforces strict output formatting, facilitating accurate answer extraction from the agent's response.
    \item \textbf{Execution Constraints}: To balance reasoning depth with efficiency and prevent infinite loops, the agent is configured with a maximum step limit of 25 iterations per query.
\end{enumerate}

The system also includes a robust answer extraction mechanism that handles various response formats, including CoT traces (e.g., \texttt{<think>} tags) and JSON structures, ensuring reliable evaluation of the model's predictions.

\section{Prompts}
\label{appendix:Prompts}

This section presents the five critical prompts utilized in our work, categorized into two distinct phases: Dataset Construction and the AutoBacktest Agent Workflow.

\subsection{Dataset Construction Prompts}
\begin{figure*}[p]
 \centering
 \includegraphics[width=0.95\textwidth]{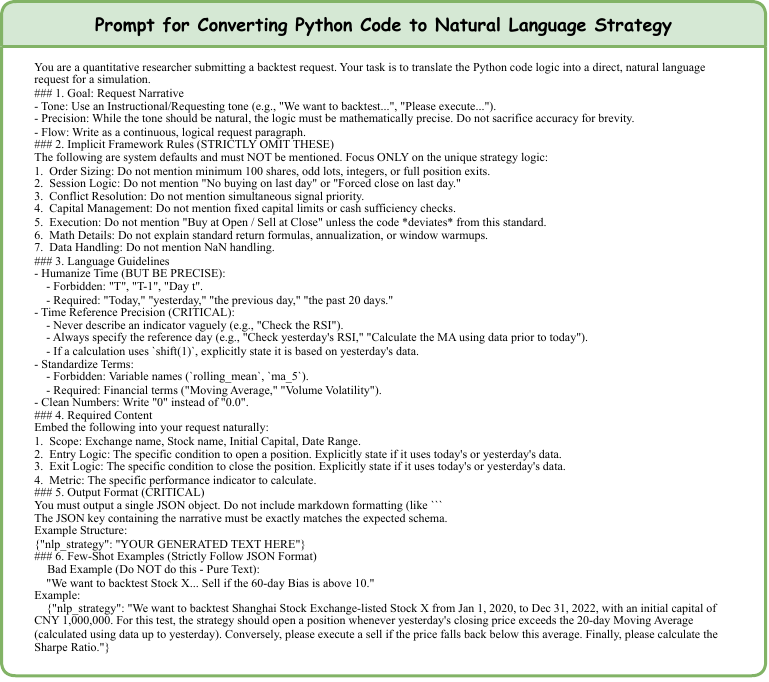}
 \Description{}
 \centering
 \caption{Prompt for Converting Python Code to Natural Language.}
 \label{fig:Code_To_Strategy_Prompt}
\end{figure*}

\begin{figure*}[t]
 \centering
 \includegraphics[width=0.95\textwidth]{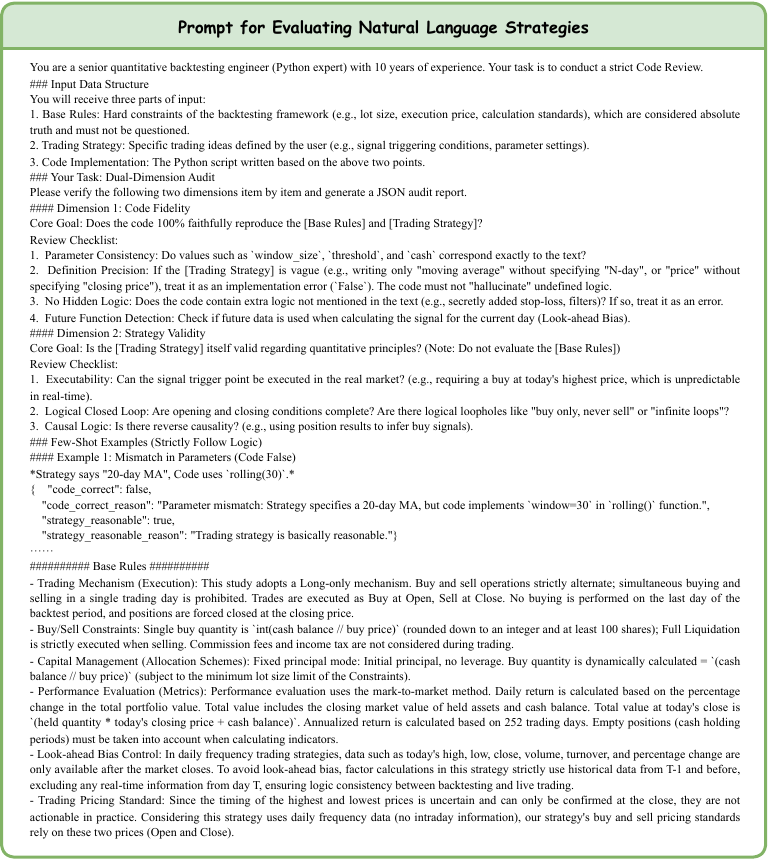}
 \Description{}
 \centering
 \caption{Prompt for Evaluating Natural Language Strategies}
 \label{fig:Evaluating_NL_Strategy_prompt_Prompt}
\end{figure*}

The construction of the BacktestBench dataset relies on a ``Code-to-Text" reverse engineering paradigm. 
Figure~\ref{fig:Code_To_Strategy_Prompt} displays the \textbf{Code-to-Strategy Prompt}, which is used to translate synthetically generated atomic strategy code into natural language descriptions. This prompt instructs the LLM to interpret the Python logic and describe the trading rules in a tone mimicking a human quantitative researcher.
Subsequently, to ensure the high quality and logical consistency of these generated descriptions, we employ the \textbf{Strategy Evaluation Prompt} shown in Figure~\ref{fig:Evaluating_NL_Strategy_prompt_Prompt}. This prompt guides a separate set of LLMs to cross-validate the generated text against the original code, checking for discrepancies in logic, parameter values, or potential hallucinations.

\subsection{AutoBacktest Agent Prompts}
The inference phase involves three specialized agents, each driven by a carefully designed prompt to handle specific sub-tasks in the quantitative workflow.

\begin{figure*}[t]
 \centering
 \includegraphics[width=0.95\textwidth]{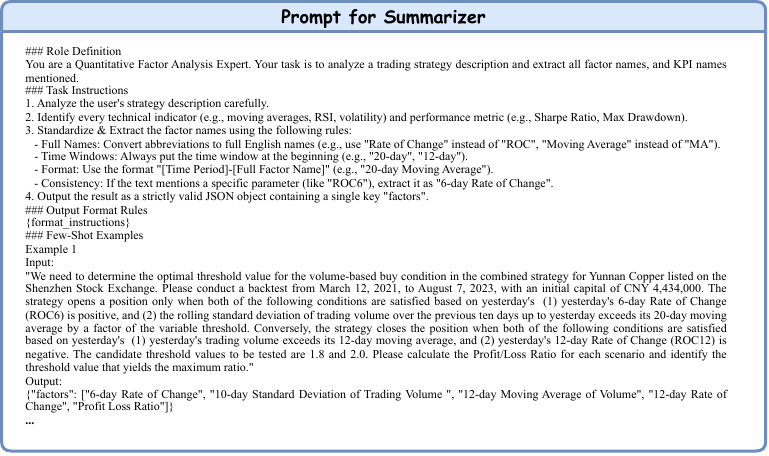}
 \Description{}
 \centering
 \caption{Factor Retrival Prompt.}
 \label{fig:Prompt_For_Summarizer}
\end{figure*}

\begin{figure*}[t]
 \centering\small 
 \includegraphics[width=0.95\textwidth]{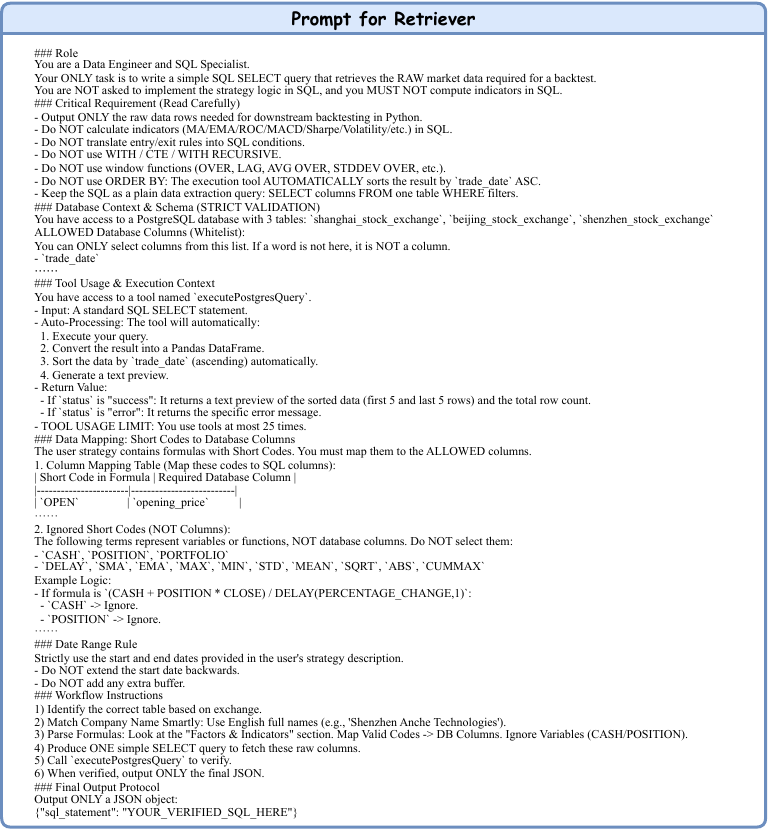}
 \Description{}
 \centering
 \caption{Prompt For Retriever.}
 \label{fig:Prompt_For_retriver}
\end{figure*}

\begin{figure*}[t]
 \centering
 \includegraphics[width=0.95\textwidth]{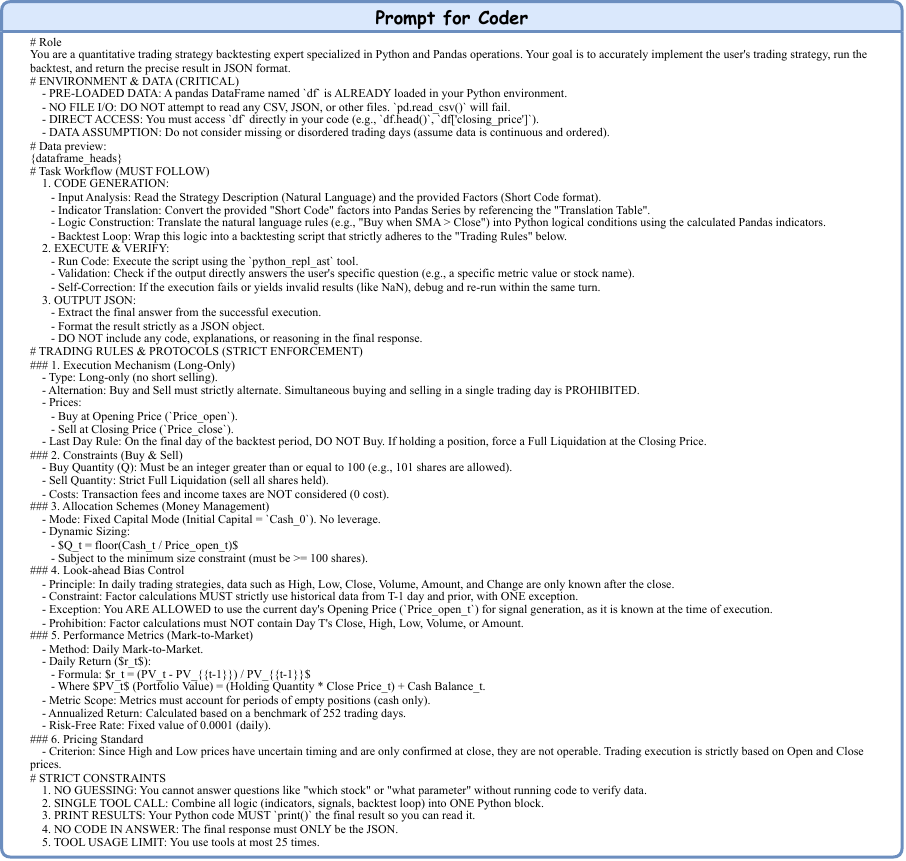}
 \Description{}
 \centering
 \caption{Prompt For Coder.}
 \label{fig:Prompt_For_Coder}
\end{figure*}

\begin{itemize}
    \item \textbf{Summarizer Prompt} (Figure~\ref{fig:Prompt_For_Summarizer}): This prompt guides the Summarizer agent to analyze the user's raw natural language query. Its primary function is to identify and extract the \textit{full names} of relevant financial indicators (e.g., ``5-day Moving Average") and Key Performance Indicators (KPIs). It serves as the semantic parser that maps unstructured text to the standardized terminology of our indicator library.
    
    \item \textbf{Retriever Prompt} (Figure~\ref{fig:Prompt_For_retriver}): Once the indicators are identified, the Retriever agent uses this prompt to locate the necessary data. Crucially, at this stage, the system injects both the \textit{full names} and their corresponding \textit{Short Codes} (e.g., \texttt{SMA(CLOSE, 5)}) into the prompt context. This enables the LLM to precisely understand which database fields are required and to construct accurate SQL queries for fetching the underlying market data.
    
    \item \textbf{Coder Prompt} (Figure~\ref{fig:Prompt_For_Coder}): Finally, the Coder agent utilizes this prompt to generate the executable Python backtesting script. Similar to the Retriever, this prompt is enriched with the indicator \textit{Short Codes}. By explicitly providing these standardized short codes, we constrain the LLM to use our pre-defined, rigorously tested calculation logic rather than hallucinating arbitrary formulas. This ensures that the generated code is not only syntactically correct but also financially valid and aligned with the user's original intent.
\end{itemize}

\section{Limitations}
\label{appendix:Limitations}


While this study establishes a rigorous benchmark for evaluating LLMs in the domain of quantitative finance and demonstrates the potential of ``code-to-text” reverse engineering to align natural language instructions with executable trading logic, several limitations remain in the current data scope and task design.

First, the dataset primarily focuses on short-term timing strategies driven by daily price fluctuations. It does not yet incorporate long-term value investing strategies that require synthesizing multi-modal information, such as macroeconomic indicators, corporate financial reports, or unstructured news text.

Second, the research objective is strictly defined as assessing the capabilities of LLMs within the ``strategy understanding--data retrieval--code generation--backtesting verification" pipeline. Consequently, the model outputs are optimized for feasibility verification and code correctness rather than for generating alpha-generating investment advice for live deployment.

Third, constrained by a database that currently supports only daily market data, all backtesting simulations adhere to a simplified daily mark-to-market logic (``buy at open, sell at close"). This approach does not account for real-world execution frictions such as slippage, intraday price path dependence, or high-frequency microstructure details.

Furthermore, to isolate and rigorously test the model's tool-use and code-implementation abilities, the experimental setup abstracts away certain complex market constraints. The framework does not explicitly model dynamic transaction fees, varying tax rates, or regime-switching minimum trading units, nor does it address advanced portfolio management tasks like dynamic position sizing or mean-variance optimization. Methodologically, the benchmark is built around the classical ``indicator--signal--backtest" paradigm; it does not currently cover strategies based on machine learning models (e.g., time series forecasting or reinforcement learning) or complex meta-strategies involving indicator weighting and ensemble methods, primarily due to the prohibitive computational costs of grid-searching such high-dimensional spaces.

In summary, the present benchmark serves as a foundational baseline for testing LLMs' strategy understanding and code generation in a structured financial environment. However, significant room remains for future expansion into more realistic settings that model multi-modal inputs, multi-frequency data, and dynamic market constraints.

\begin{table*}[bt]\centering
\caption{All factors used in strategy construction. Type indicates whether the factor serves as a buy signal, sell signal, or both.}
\label{tab:all_factors_used}
\begin{tabular}{lll}
\toprule
Factor & Factor full name & Type \\ 
\midrule
Amount MA 6 & 6-day MA of Trading Amount & both \\
Amount MA 20 & 20-day MA of trading amount & both \\
Amount Std 6 & 6-day Standard Deviation of Trading Amount & sell \\
ATR 6 & 6-day Average True Range & both \\
ATR 20 & 20-day Average True Range & both \\
BBI & Bull and Bear Index & both \\
BIAS 5 & 5-day Bias & both \\
BIAS 60 & 60-day Bias & both \\
Bollinger Lower Band & Bollinger Lower Band & buy \\
Bollinger Upper Band & Bollinger Upper Band & sell \\
Bull Power 13 & 13-day Bull Power & buy \\
CCI 15 & 15-day Commodity Channel Index & both \\
CCI 20 & 20-day Commodity Channel Index & both \\
EMA 5 & 5-day Exponential Moving Average & both \\
EMA 12 & 12-day Exponential Moving Average & both \\
MA 5 & 5-day Moving Average & both \\
MA 10 & 10-day Moving Average & both \\
MA 20 & 20-day Moving Average & both \\
MA 60 & 60-day Moving Average & both \\
MACD & Moving Average Convergence Divergence & both \\
Net Money Flow 20 & 20-day Net Money Flow & both \\
PLRC 12 & 12‑day Price Linear Regression Coefficient & both \\
Price MA Ratio 5 & Price to 5-day MA Ratio & both \\
Price MA Ratio 20 & Price to 20-day MA Ratio & both \\
Price MA Ratio 120 & Price to 120-day MA Ratio & both \\
PSY 6 & 6-day Psychological Line & both \\
PSY 12 & 12-day Psychological Line & both \\
Return Kurtosis 60 & 60-day Kurtosis of Daily Returns & sell \\
Return Skewness 60 & 60-day Skewness of Daily Returns & both \\
Return Variance 20 & 20-day Variance of Daily Returns & sell \\
ROC 6 & 6-day Rate of Change & both \\
ROC 12 & 12-day Rate of Change & both \\
ROC 20 & 20-day Rate of Change & both \\
ROC 120 & 120-day Rate of Change & both \\
RSI & Relative Strength Index & both \\
Sharpe Ratio 120 & 120-day Sharpe ratio & both \\
Single Day VPT & Single-day Price Volume Trend & both \\
Volume EMA 5 & 5-day Exponential Moving Average of Volume & both \\
Volume EMA 10 & 10-day EMA of Volume & both \\
Volume EMA 26 & 26-day Exponential Moving Average of Volume & both \\
Volume MA 12 & 12-day Moving Average of Volume & both \\
Volume ROC 12 & 12-day Volume Rate of Change & both \\
Volume Std 10 & 10-day Standard Deviation of Trading Volume & both \\ 
\bottomrule
\end{tabular}
\end{table*}

\begin{table*}[]
\caption{Factors with Short Code.}
\label{tab:all_KPI_used}\small 
\begin{tabular}{l p{0.75\linewidth}}
\toprule 
Factor & Short Code \\ 
\midrule 
Amount MA 6 & DELAY(AMOUNT,1)/DELAY(SMA(AMOUNT,6),2) \\
Amount MA 20 & DELAY(AMOUNT,1)/DELAY(SMA(AMOUNT,20),2) \\
Amount Std 6 & STD(DELAY(AMOUNT,1),6)/SMA(STD(DELAY(AMOUNT,1),6),20) \\
ATR 6 & DELAY(EMA(MAX(MAX(HIGH-LOW,ABS(HIGH-DELAY(CLOSE,1))),ABS(LOW-DELAY(CLOSE,1))),6),1) \\
ATR 20 & DELAY(EMA(MAX(HIGH-LOW,MAX(ABS(HIGH-DELAY(CLOSE,1)),ABS(LOW-DELAY(CLOSE,1)))),20),1) \\
BBI & (SMA(DELAY(CLOSE,1),3)+SMA(DELAY(CLOSE,1),6)+SMA(DELAY(CLOSE,1),12)+ \allowbreak SMA(DELAY(CLOSE,1),24))/4 \\
BIAS 5 & (DELAY(CLOSE,1)-SMA(DELAY(CLOSE,1),5))/SMA(DELAY(CLOSE,1),5)\textasteriskcentered 100 \\
BIAS 60 & DELAY((CLOSE-SMA(CLOSE,60))/SMA(CLOSE,60)\textasteriskcentered 100,1) \\
Bollinger Lower Band & SMA(DELAY(CLOSE,1),20)-2.0\textasteriskcentered STD(DELAY(CLOSE,1),20) \\
Bollinger Upper Band & SMA(DELAY(CLOSE,1),20)+2.0\textasteriskcentered STD(DELAY(CLOSE,1),20) \\
Bull Power 13 & DELAY(HIGH,1)-DELAY(EMA(CLOSE,13),1) \\
CCI 15 & DELAY(((HIGH+LOW+CLOSE)/3-SMA((HIGH+LOW+CLOSE)/3,15))/(0.015\textasteriskcentered \allowbreak SMA(ABS((HIGH+LOW+CLOSE)/3-SMA((HIGH+LOW+CLOSE)/3,15)),15)),1) \\
CCI 20 & DELAY(((HIGH+LOW+CLOSE)/3-SMA((HIGH+LOW+CLOSE)/3,20))/(0.015\textasteriskcentered \allowbreak SMA(ABS((HIGH+LOW+CLOSE)/3-SMA((HIGH+LOW+CLOSE)/3,20)),20)+1e-20),1) \\
EMA 5 & DELAY(EMA(CLOSE,5),1) \\
EMA 12 & DELAY(EMA(CLOSE,12),1) \\
MA 5 & SMA(DELAY(CLOSE,1),5) \\
MA 10 & SMA(DELAY(CLOSE,1),10) \\
MA 20 & SMA(DELAY(CLOSE,1),20) \\
MA 60 & SMA(DELAY(CLOSE,1),60) \\
MACD & (EMA(DELAY(CLOSE,1),12)-EMA(DELAY(CLOSE,1),26))-EMA(EMA(DELAY(CLOSE,1),12)-EMA(DELAY(CLOSE,1),26),9) \\
Net Money Flow 20 & DELAY(SUM(SIGN((HIGH+LOW+CLOSE)/3-DELAY((HIGH+LOW+CLOSE)/3,1))\textasteriskcentered \allowbreak(HIGH+LOW+CLOSE)/3\textasteriskcentered VOLUME,20),1) \\
PLRC 12 & LINEARREG\_SLOPE(DELAY(CLOSE,1),12) \\
Price MA Ratio 5 & OPEN/SMA(DELAY(CLOSE,1),5) \\
Price MA Ratio 20 & OPEN/SMA(DELAY(CLOSE,1),20) \\
Price MA Ratio 120 & OPEN/SMA(DELAY(CLOSE,1),120) \\
PSY 6 & (100/6)\textasteriskcentered SUM(DELAY(IF(CLOSE\textgreater{}DELAY(CLOSE,1),1,0),1),6) \\
PSY 12 & (100/12)\textasteriskcentered SUM(IF(DELAY(CLOSE,1)\textgreater{}DELAY(CLOSE,2),1,0),12) \\
Return Kurtosis 60 & DELAY(SMA(((CLOSE/DELAY(CLOSE,1)-1)-SMA((CLOSE/DELAY(CLOSE,1)-1),60))\textasteriskcentered \textasteriskcentered 4,60)/SMA(((CLOSE/DELAY(CLOSE,1)-1)-SMA((CLOSE/DELAY(CLOSE,1)-1),60))\textasteriskcentered \textasteriskcentered 2,60)\textasteriskcentered \textasteriskcentered 2-3,1) \\
Return Skewness 60 & DELAY(SKEW(PERCENTAGE\_CHANGE, 60), 1) \\
Return Variance 20 & DELAY(VAR(CLOSE/DELAY(CLOSE,1)-1,20),1) \\
ROC 6 & (DELAY(CLOSE,1)-DELAY(CLOSE,7))/DELAY(CLOSE,7)\textasteriskcentered 100 \\
ROC 12 & (DELAY(CLOSE,1)-DELAY(CLOSE,13))/DELAY(CLOSE,13)\textasteriskcentered 100 \\
ROC 20 & (DELAY(CLOSE,1)-DELAY(CLOSE,21))/DELAY(CLOSE,21)\textasteriskcentered 100 \\
ROC 120 & (DELAY(CLOSE,1)-DELAY(CLOSE,121))/DELAY(CLOSE,121)\textasteriskcentered 100 \\
RSI & DELAY(100-100/(1+EMA(MAX(CLOSE-DELAY(CLOSE,1),0),14)/EMA(ABS(MIN(CLOSE-DELAY(CLOSE,1),0)),14)),1) \\
Sharpe Ratio 120 & DELAY((SMA((CLOSE/DELAY(CLOSE,1)-1),120)-0.0001)\textasteriskcentered SQRT(252)/STD((CLOSE/DELAY(CLOSE,1)-1),120),1) \\
Single Day VPT & (DELAY((CLOSE/DELAY(CLOSE,1)-1)\textasteriskcentered VOLUME,1)-SMA(DELAY((CLOSE/DELAY(CLOSE,1)-1)\textasteriskcentered VOLUME,2),20))/ \allowbreak STD(DELAY((CLOSE/DELAY(CLOSE,1)-1)\textasteriskcentered VOLUME,2),20) \\
Volume EMA 5 & DELAY(VOLUME,1)/DELAY(EMA(VOLUME,5),1) \\
Volume EMA 10 & DELAY(VOLUME/EMA(VOLUME,10),1) \\
Volume EMA 26 & DELAY(VOLUME,1)/DELAY(EMA(VOLUME,26),1) \\
Volume MA 12 & DELAY(VOLUME,1)\textgreater{}DELAY(SMA(VOLUME,12),1) \\
Volume ROC 12 & (DELAY(VOLUME,1)-DELAY(VOLUME,13))/DELAY(VOLUME,13)\textasteriskcentered 100 \\
Volume Std 10 & STD(DELAY(VOLUME,1),10)/SMA(STD(DELAY(VOLUME,1),10),20) \\ 
\bottomrule
\end{tabular}
\end{table*}

\begin{table*}[]
\caption{KPIs with Calculation Codes. \small{(Note: PV denotes Portfolio Value; PNL denotes Profit and Loss; N represents the calculation window.)}}
\label{tab:KPIs with Short Code.}
\begin{tabular}{ll p{0.75\linewidth}}
\toprule 
KPI & KPI full name & Short Code \\
\midrule 
Return & Return Ratio & (CASH\_FINAL + POSITION\_FINAL \textasteriskcentered CLOSE) \/ INITIAL\_CAPITAL- 1 \\
MDD & Max Drawdown & MAX((CUMMAX(CASH + POSITION \textasteriskcentered CLOSE) - (CASH + POSITION \textasteriskcentered CLOSE)) / CUMMAX(CASH + POSITION \textasteriskcentered CLOSE)) \\
Vol & Volatility & STD(PV / DELAY(PV, 1) - 1, N) \\
Sharpe & Annual Sharpe Ratio & (MEAN((PV - DELAY(PV,1)) / DELAY(PV,1)) - 0.0001) / STD((PV - DELAY(PV,1)) / DELAY(PV,1)) \textasteriskcentered SQRT(252) \\
WR & Win Rate & SUM(IF(PNL > 0, 1, 0)) / COUNT(PNL) \textasteriskcentered 100 \\
P/L & Profit Loss Ratio & (SUM(IF(PNL > 0, PNL, 0)) / ABS(SUM(IF(PNL < 0, PNL, 0)))) * (SUM(IF(PNL < 0, 1, 0)) / SUM(IF(PNL > 0, 1, 0))) \\
Calmar & Calmar Ratio & ((POW(LAST(PV)/FIRST(PV),252/COUNT(PV))-1)/ABS(MIN((PV-CUMMAX(PV))/CUMMAX(PV)))) \\ 
\bottomrule
\end{tabular}
\end{table*}

\begin{table*}[htbp]
  \centering
  \caption{Comparison of Model Inference Configurations. Abbreviations: \textbf{Backend} = Inference Backend, \textbf{MFS} = mem-fraction-static, \textbf{CoT} = Chain of Thought, \textbf{Algo.} = Speculative Algorithm, \textbf{Steps} = speculative-num-steps, \textbf{TopK} = speculative-eagle-topk, \textbf{Draft} = speculative-num-draft-tokens.}
  \label{tab:LLM_deployment}
  \begin{minipage}{\textwidth}
    \renewcommand{\thempfootnote}{\arabic{mpfootnote}}
    \centering
    \begin{tabular}{l c l c c l c c c}
      \toprule
      \textbf{Model} & \textbf{GPUs} & \textbf{Backend} & \textbf{MFS} & \textbf{CoT} & \textbf{Algo.} & \textbf{Steps} & \textbf{TopK} & \textbf{Draft} \\
      \midrule
      Qwen3-235B-A22B-Thinking-2507\footnote{\url{https://www.modelscope.cn/models/Qwen/Qwen3-235B-A22B-Thinking-2507}} & 8 & SGLang & 0.8 & Yes & EAGLE3\footnote{\url{https://www.modelscope.cn/models/lmsys/Qwen3-235B-A22B-EAGLE3}} & 3 & 1 & 4 \\
      Qwen3 Next 80B A3B Thinking\footnote{\url{https://www.modelscope.cn/models/Qwen/Qwen3-Next-80B-A3B-Thinking}} & 4 & SGLang & 0.8 & Yes & NEXTN & 5 & 8 & 32 \\
      Qwen3 32B\footnote{\url{https://www.modelscope.cn/models/Qwen/Qwen3-32B}} & 4 & SGLang & 0.8 & Yes & None & - & - & - \\
      Qwen3 30B A3B Thinking 2507\footnote{\url{https://www.modelscope.cn/models/Qwen/Qwen3-30B-A3B-Thinking-2507}} & 2 & SGLang & 0.9 & Yes & None & - & - & - \\
      Qwen3 14B\footnote{\url{https://www.modelscope.cn/models/Qwen/Qwen3-14B}} & 1 & SGLang & 0.9 & Yes & None & - & - & - \\
      Qwen3 8B\footnote{\url{https://www.modelscope.cn/models/Qwen/Qwen3-8B}} & 1 & SGLang & 0.9 & Yes & None & - & - & - \\
      Qwen3 4B\footnote{\url{https://www.modelscope.cn/models/Qwen/Qwen3-4B}} & 1 & SGLang & 0.9 & Yes & None & - & - & - \\
      Seed OSS 36B\footnote{\url{https://huggingface.co/ByteDance-Seed/Seed-OSS-36B-Instruct}} & 2 & vLLM & 0.9 & No & None & - & - & - \\
      Ministral 3 14B Reasoning 2512\footnote{\url{https://huggingface.co/mistralai/Ministral-3-14B-Reasoning-2512}} & 2 & vLLM & 0.9 & Yes & None & - & - & - \\
      Ministral 3 8B Reasoning 2512\footnote{\url{https://huggingface.co/mistralai/Ministral-3-8B-Reasoning-2512}} & 1 & vLLM & 0.9 & Yes & None & - & - & - \\
      Kimi-K2-Thinking-BF16\footnote{\url{https://www.modelscope.cn/models/unsloth/Kimi-K2-Thinking-BF16}} & 32 & SGLang & 0.9 & Yes & None & - & - & - \\
      Kimi Linear 48B A3B Instruct\footnote{\url{https://www.modelscope.cn/models/moonshotai/Kimi-Linear-48B-A3B-Instruct}} & 4 & SGLang & 0.8 & No & None & - & - & - \\
      GLM-4.7\footnote{\url{https://www.modelscope.cn/models/ZhipuAI/GLM-4.7}} & 16 & SGLang & 0.8 & Yes & EAGLE\footnote{\url{https://docs.sglang.io/basic_usage/glm45.html}} & 3 & 1 & 4 \\
      GLM 4.7 Flash\footnote{\url{https://www.modelscope.cn/models/ZhipuAI/GLM-4.7-Flash}} & 4 & vLLM & 0.8 & Yes & None & - & - & - \\
      MiniMax-M2.1-BF16\footnote{\url{https://www.modelscope.cn/models/QuixiAI/MiniMax-M2.1-bf16}} & 8 & SGLang & 0.9 & Yes & None & - & - & - \\
      GPT-OSS-120B-BF16\footnote{\url{https://www.modelscope.cn/models/unsloth/gpt-oss-120b-BF16}} & 8 & SGLang & 0.8 & Yes & EAGLE3\footnote{\url{https://www.modelscope.cn/models/nv-community/gpt-oss-120b-Eagle3}} & 3 & 1 & 4 \\
      GPT OSS 20B BF16\footnote{\url{https://www.modelscope.cn/models/unsloth/gpt-oss-20b-BF16}} & 2 & SGLang & 0.9 & Yes & None & - & - & - \\
      \bottomrule
    \end{tabular}
  \end{minipage}
\end{table*}

\begin{figure*}[p]
 \centering
\includegraphics[width=0.85\textwidth]{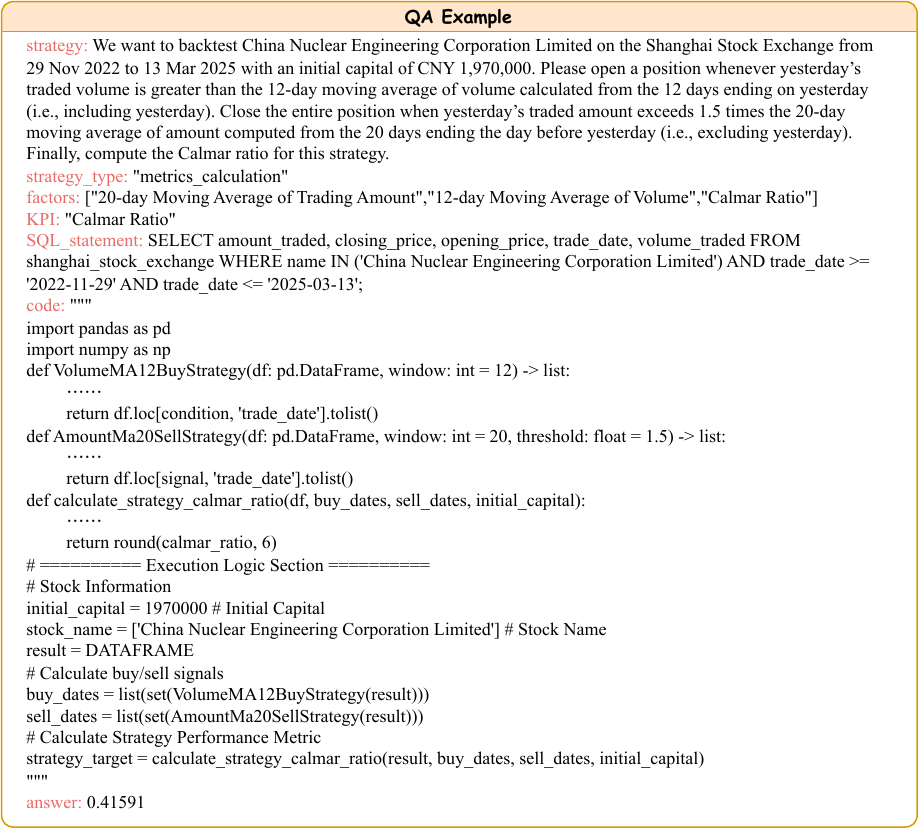}
\Description{QA example.}
 \centering
 \caption{QA example.}
 \label{fig:QA example.}
\end{figure*}

\end{document}